\documentclass[conference]{IEEEtran}
\IEEEoverridecommandlockouts
\usepackage{graphicx} 
\usepackage{amsfonts}
\usepackage{amsmath}
\usepackage{amssymb}
\usepackage{amsthm}
\usepackage{hyperref}
\usepackage{url}            
\usepackage{booktabs}       
\usepackage{nicefrac}       
\usepackage{microtype}      
\usepackage{xcolor}         

\usepackage{paralist}
\newtheorem{definition}{Definition}
\usepackage{algorithm}
\usepackage{algpseudocode}
\usepackage{tabularx}
\usepackage{rotating}
\usepackage{graphicx}
\usepackage{subfig}

\usepackage{afterpage}
\usepackage{multirow} 
\def\BibTeX{{\rm B\kern-.05em{\sc i\kern-.025em b}\kern-.08em
    T\kern-.1667em\lower.7ex\hbox{E}\kern-.125emX}}

\newif\ifbold
\boldfalse

\usepackage[square,numbers]{natbib}
\title{MOB-GCN: A Novel Multiscale Object-Based Graph Neural Network for Hyperspectral Image Classification}
\author{
\IEEEauthorblockN{Tuan-Anh Yang\thanks{This work is done during Tuan-Anh Yang's internship under the supervision of Dr. Truong-Son Hy in the Department of Computer Science, Univerity of Alabama at Birmingham, United States.}\thanks{Tuan-Anh Yang is with the Faculty of Information Technology, University of Science, VNU-HCM, Vietnam (e-mail: ytanh21@apcs.fitus.edu.vn)}}
\IEEEauthorblockA{
\textit{VNU-HCM University of Science}\\
ytanh21@apcs.fitus.edu.vn}
\and
\IEEEauthorblockN{Truong-Son Hy\thanks{Truong-Son Hy is with the Department of Computer Science, Univerity of Alabama at Birmingham, United States (e-mail: thy@uab.edu) }}
\IEEEauthorblockA{
\textit{University of Alabama at Birmingham}\\
thy@uab.edu}
\and
\IEEEauthorblockN{Phuong D. Dao\thanks{Phuong D. Dao is with the Department of Agricultural Biology, Colorado State University, United States (e-mail: phuong.dao@colostate.edu)}}
\IEEEauthorblockA{
\textit{Colorado State University}\\
phuong.dao@colostate.edu}
}
\date{October 2024}

\begin{document}

\maketitle

\begin{abstract}
This paper introduces a novel multiscale object-based graph neural network called MOB-GCN for hyperspectral image (HSI) classification. The central aim of this study is to enhance feature extraction and classification performance by utilizing multiscale object-based image analysis (OBIA). Traditional pixel-based methods often suffer from low accuracy and speckle noise, while single-scale OBIA approaches may overlook crucial information of image objects at different levels of detail. MOB-GCN addresses this issue by extracting and integrating features from multiple segmentation scales to improve classification results using the Multiresolution Graph Network (MGN) architecture that can model fine-grained and global spatial patterns. By constructing a dynamic multiscale graph hierarchy, MOB-GCN offers a more comprehensive understanding of the intricate details and global context of HSIs. Experimental results demonstrate that MOB-GCN consistently outperforms single-scale graph convolutional networks (GCNs) in terms of classification accuracy, computational efficiency, and noise reduction, particularly when labeled data is limited. The implementation of MOB-GCN is publicly available at \url{https://github.com/HySonLab/MultiscaleHSI}.  
\end{abstract}


\section{Introduction} \label{sec:introduction}

Hyperspectral images (HSIs) provide rich spectral information, making them valuable for various applications. \citep{lu2020} This unique characteristic has led to the widespread use of HSI in various applications, such as classification \citep{melgani2004, fang2015, fang2018, DAO2021102542}, object tracking \citep{wang2010, uzkent2016, uzkent2017} and object detection \citep{pan2003, liu2016, zhang2016}, environmental monitoring \citep{ellis2004, manfreda2018}, and vegetation health assessment \citep{DAO2021102364}. However, classification of HSIs remains challenging due to the limited availability of labeled training data, and the high dimensionality of the data, significant spatial variability. Traditional pixel-based classification methods often suffer from low accuracy and speckle noise. To tackle these issues, object-based image analysis (OBIA) has emerged as a promising image interpretation approach as it reduces noise in the classified map and improves classification accuracy and computational efficiency \citep{20-photo-j, blaschke2010object}. Object-Based Image Analysis (OBIA) involves two main steps: image segmentation, which clusters pixels into meaningful objects, and image classification, which categorizes these objects into specific classes. Between the two steps, determining the optimal segmentation scales and extracting object features are critical to achieving high-quality classification outcomes \citep{20-photo-j, draguct2010esp}. Previous studies are limited to utilizing features from a single segmentation scale \citep{20-photo-j}, which can overlook important hierarchical relationships within the data.

To overcome this limitation, \citet{Hy_2023} introduced Multiresolution Graph Networks (MGN), which dynamically construct multiple resolutions of the input graph using data-driven clustering. MGN employs two GNN modules at each resolution: one for graph representation learning and another for graph coarsening. This adaptive clustering process is crucial to capture both local and global features, allowing the model to gradually focus on broader structures as necessary.

Motivated by the flexibility and efficiency of MGN, we propose a novel Multiscale Object-Based Graph Convolutional Network (MOB-GCN) that incorporates a multiresolution mechanism for robust HSI classification. Our approach utilizes the MGN architecture and provides an ablation study on benchmark datasets. In summary, our contributions are:
\begin{enumerate} 
    \item Developing an automatic optimal segmentation scale determination method for effective extraction of image object's spatial and spectral features from hyperspectral images at at different scales;
    \item Developing a novel multiscale object-based classification method that integrates features extracted from multiple segmentation scales to improve the overall classification accuracy;
    \item Comparing the performance of our proposed MOB-GCN model with the single-scale GCN model to demonstrate the advantages of our multiscale method.
\end{enumerate}

This approach integrates multiresolution graph learning into a unified framework, providing a more comprehensive understanding of hyperspectral images compared to single-scale methods.


\section{Related Work} \label{sec:related_work}

\paragraph{Graph Representation Learning} Graph representation learning is essential for leveraging structural information in graph-structured data by embedding it in low-dimensional spaces. \cite{deepwalk, node2vec, network_embedding, pmlr-v196-hy22a} Methods have evolved from early spectral techniques to sophisticated approaches such as graph neural networks (GNNs). Recent advances, such as graph transformers \cite{NEURIPS2019_9d63484a, kim2022pure, pmlr-v202-cai23b, 10.1063/5.0152833} and graph attention networks (GATs) \cite{velickovic2018graph}, adapt transformers and attention mechanisms to graph domains to capture long-range node interactions. These innovations enhance the processing and analysis of graph-structured data.


\paragraph{Multiscale Graph Methods} Multiscale graph methods effectively capture hierarchical structures and integrate local and global information. They are beneficial in domains with multiresolution characteristics like hyperspectral imaging. Multiresolution graph learning dynamically constructs hierarchical graph representations through graph coarsening. Previous works have proposed multiresolution Graph Neural Networks \cite{Hy_2023, pmlr-v184-hy22a, 10.1063/5.0152833, trang2024scalable} and Graph Transformers that use data-driven clustering to partition graphs into multiple levels. In hyperspectral imaging, multiscale methods are relatively underexplored, despite using superpixels for classification has shown promise. However, these methods often rely on a single resolution or scale, which may overlook hierarchical relationships crucial for robust classification.

\paragraph{Hyperspectral Image Classification} 
The semi-supervised classification of hyperspectral images (HSIs) has been a focal point of research within the remote sensing community, with graph-based learning techniques emerging as a prominent approach. In these methods, data points are represented as nodes, while edges and weights encode the similarity between them, enabling effective spatial-spectral modeling.
One of the pioneering methods in this area was introduced by \citet{campsvalls2007}, where a combination of spectral and spatial kernels, along with the Nystr\"om extension for matrix approximation, was used for HSI classification. However, this method exhibited relatively low accuracy compared to more recent techniques. Later, \citet{gao2014} improved the performance by introducing a bilayer graph-based learning algorithm. Their approach combined a pixel-based graph, similar to \cite{campsvalls2007}, with a hypergraph constructed from grouping relations derived through unsupervised learning. 

\paragraph{Object-Based Image Analysis}
OBIA techniques cluster pixels with similar spectral and textural characteristics to create image objects that more accurately represent real-world surface features. By classifying a smaller number of image objects, the OBIA approaches are more computationally efficient compared to pixel-based methods. Previous studies have proposed methods for determining optimal segmentation scales but are limited to utilizing features from a single segmentation scale in classification. In addition to the pixel's grey value, more features can be extracted and be included in the classification step to improve accuracy in the OBIA approaches \citep{20-photo-j, dao2019, dao2015}. By grouping pixels to form image objects, the method reduces the speckle noise effect in the classified image and generates more meaningful classification results \citep{he2015}. To achieve accurate and robust classification results in OBIA approaches, it is critical to determine optimal segmentation scales and extract neccessary features for classification models. Previous studies \citep{20-photo-j, draguct2010esp, sellars2020, wan2019multiscaledynamicgraphconvolutional, zhang2021hyperspectralimagesegmentationbased} have proposed several optimal scale selection methods and achieved promissing results. However, these studies are limited to utilizing features from a single segmentation scale in classification, and no study has incorporated information extracted from multiple optimal scales to improve classification outcomes.

Our work addresses these issues by integrating multiresolution graph learning with object-based image analysis into a unified framework. Unlike traditional single-scale GNNs or static superpixel-based methods, we employ Felzenszwalb’s superpixel segmentation and construct a dynamic multiscale graph hierarchy to model both fine-grained and global spatial patterns.

\section{Method} \label{sec:method}

\begin{figure*}[h!]
    \centering
    \includegraphics[width=\linewidth]{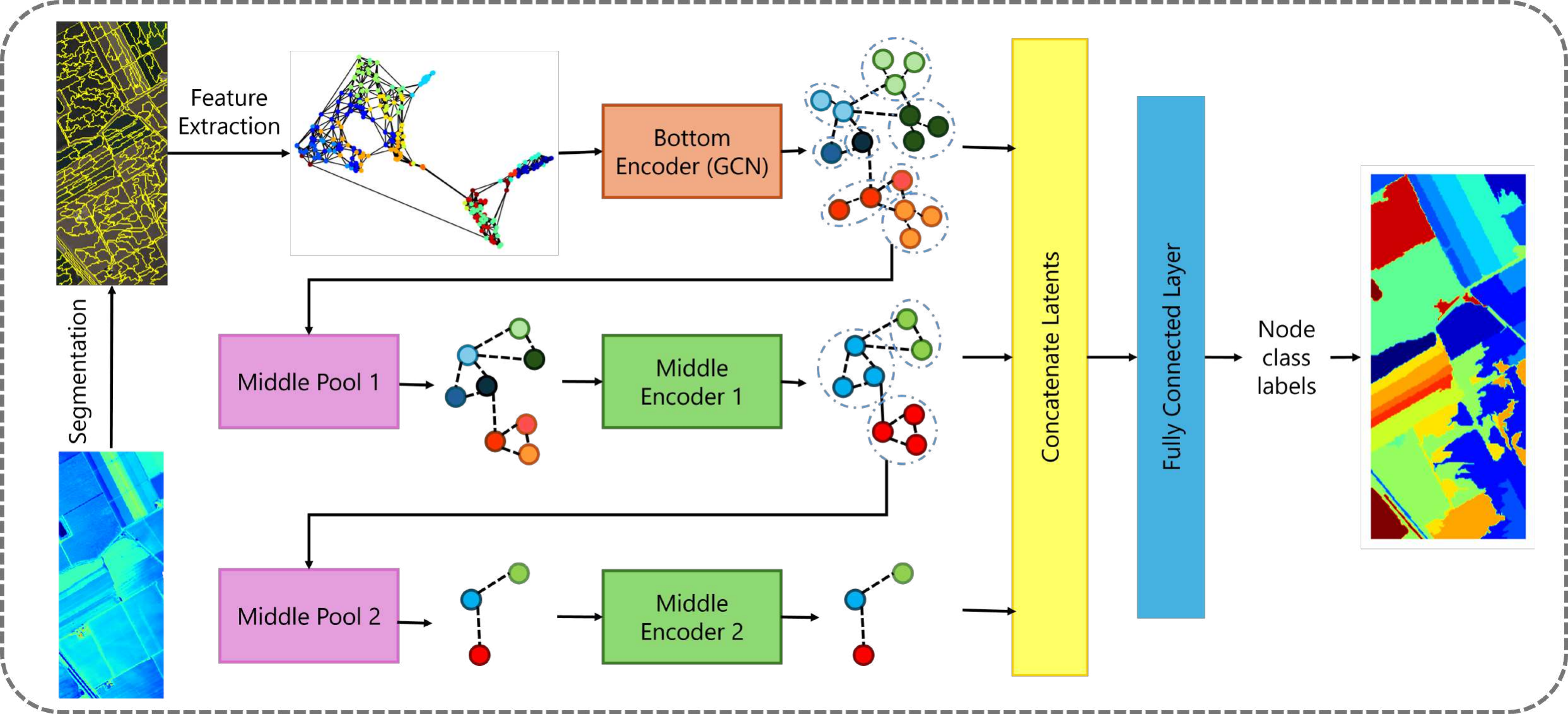}
    \caption{The proposed framework follows a structured pipeline for hyperspectral image (HSI) classification. First, the HSI is read and undergoes dimensionality reduction before applying superpixel segmentation. Features are then extracted from each superpixel and, along with the initial labeling, are used to construct a superpixel-based graph. This graph is first processed by a bottom encoder before undergoing recursive pooling and encoding at multiple resolutions. The latent representations from all resolutions, including the bottom encoding, are concatenated and passed into a final classifier. The predicted labels are then mapped back to the superpixel regions, producing the final classification of the HSI.}
    \label{fig:overall_architecture}
\end{figure*}

In this study, we aim to achieve precise classification predictions for a vast amount of unlabeled data while relying on a minimal set of labeled samples. We formulate the classification task within the framework of semi-supervised learning (SSL).

\begin{definition}[Semi-Supervised Classification Task]  
Given a labeled dataset \(\{(x_i, y_i)\}_{i=1}^l\) and a label set \(\mathcal{L} = \{1, \ldots, c\}\) where \(\{y_i\}_{i=1}^l \in \mathcal{L}\), the goal is to learn a function \(f: \mathbb{R}^d \rightarrow \mathbb{R}^{c+u}\) that leverages the unlabeled data \(\{x_k\}_{k=l+1}^{l+u}\) to enhance prediction accuracy for \(\{x_k\}_{k=l+1}^{l+u}\).  
\end{definition}

\subsection{Superpixel Segmentation} \label{sec:segmentation}

Superpixels are perceptually meaningful, connected regions that group pixels based on similarities in color or other features, first introduced by \citet{renmalik2003}. Since then, various algorithmic approaches have been developed \citep{achanta2012, felzenszwalb2004}. Defining appropriate local regions is crucial for extracting spatial features in spectral-spatial models. While fixed-size windows (e.g., \citet{ertem2020}) have shown promising results, they constrain the ability to fully capture spatial context. In contrast, superpixels provide adaptive regions that enhance discriminative information, as demonstrated by \citet{fang2015kernals}. \citet{cui2018} further highlighted this by employing a superpixel-based random walker to refine an SVM probability map with significant success. Additionally, Cui et al. showed that superpixel spectra are more stable and less sensitive to noise than individual pixel spectra, making superpixel-based approaches more robust to image noise.


\begin{definition}[Superpixel Segmentation]  
Given an image \( I : \Lambda \to \mathbb{R}^d \), where \( \Lambda \subset \mathbb{Z}^2 \) represents the image domain, superpixel segmentation partitions \( \Lambda \) into a set of regions \(\{S_i\}_{i=1}^n\). Each superpixel \( S_i \) is defined as \( S_i = \{x \in \Lambda : f(x) = i\} \), where \( f: \Lambda \to \{1, \ldots, n\} \) is a labeling function that assigns each pixel \( x \) to one of the \( n \) superpixels based on a feature function.  
\end{definition}

We propose using the Felzenszwalb segmentation algorithm \citep{felzenszwalb2004} as an enhancement to superpixel-based methods, which predominantly rely on SLIC \citep{achanta2010slic} and its variants.

\begin{figure}[h]
    \centering
    \includegraphics[width=\linewidth]{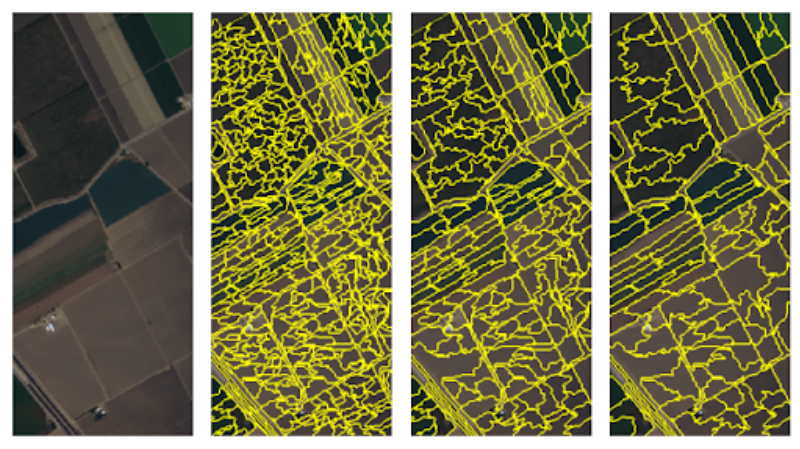}
    \caption{The Salinas HSI segmented using Felzenszwalb segmentation algorithm. \citep{felzenszwalb2004} The first figure shows a false-colored RGB image and other 3 shows the image segmented using a minimum size of 50, 100, 200 pixels respectively.}
    \label{fig:felz_salinas}
\end{figure}


\subsection{Feature Extraction} \label{sec:feature}

The next step involves extracting relevant features from the superpixels, which will be used in the subsequent graph construction step. In this work, we adopt the feature extraction approach outlined by \citet{sellars2020} for superpixels.

For each superpixel \( S_i \), we extract three distinct types of features to enhance spatial and contextual information.  

1. \textbf{Mean Feature Vector (\(\vec{S}_i^m\))}:  
   To capture localized spatial information, we apply a mean filter to each superpixel, computing the mean feature vector as:  
   \begin{equation} \label{equ:mean_filter}
       \vec{S}_i^m = \frac{\sum_{j=1}^{n_i} \hat{I}(p_{i, j})}{n_i}.
   \end{equation}  

2. \textbf{Weighted Feature Vector (\(\vec{S}_i^w\))}:  
   To incorporate spatial relationships between neighboring superpixels, we compute a weighted combination of the mean feature vectors of adjacent superpixels. Adjacency is determined using 4-connectivity (left, right, up, and down) within the image grid. For each superpixel \( S_i \), let \( \zeta_i = \{z_1, z_2, \dots, z_J\} \) denote the set of indices of its \( J \) adjacent superpixels. The weighted feature vector is then given by:  
   \begin{equation} \label{equ:weighted_filter}
       \vec{S}_i^w = \sum_{j=1}^{J} w_{i, z_j} \vec{S}_{z_j}^m,
   \end{equation}  
   where the weight \( w_{i, z_j} \) between adjacent superpixels is computed using a softmax function:  
   \begin{equation} \label{equ:weight_softmax}
       w_{i, z_j} = \frac{\exp\left(-\|\vec{S}_{z_j}^m - \vec{S}_i^m\|_2^2 / h\right)}{\sum_{j=1}^{J} \exp\left(-\|\vec{S}_{z_j}^m - \vec{S}_i^m\|_2^2 / h\right)},
   \end{equation}  
   where \( h \) is a predefined scalar parameter.  

3. \textbf{Centroid Location (\(\vec{S}_i^p\))}:  
   Finally, to encode spatial positioning, we compute the centroid location of each superpixel as:  
   \begin{equation} \label{equ:centroid}
       \vec{S}_i^p = \frac{\sum_{j=1}^{n_i} p_{i, j}}{n_i}.
   \end{equation}

\subsection{Graph-based Classification} \label{sec:graphclass}

As noted by \citet{campsvalls2007}, many graph-based algorithms involve computing and manipulating large kernel matrices that include both labeled and unlabeled data. For an image with \( n \) pixels, the corresponding graph Laplacian matrix has a size of \( n \times n \), and its inversion via singular value decomposition has a computational complexity of \( O(n^3) \), making it impractical for large-scale applications. To mitigate this issue, instead of representing each pixel as a graph node, we use superpixels as nodes, significantly reducing the node count since \( K \ll n \). This approach allows efficient matrix operations without requiring approximations while also improving classification accuracy by defining meaningful local regions within the data.

Using the extracted features and the superpixel-based node set, we construct a weighted, undirected graph \( G = (V, E, W) \). The edge weight between adjacent superpixels \( S_i \) and \( S_j \) is defined using two Gaussian kernels:  
\begin{equation} \label{equ:graph_weight}
    w_{ij} = s_{ij}l_{ij},
\end{equation}  
where the individual components are given by:  
\begin{equation} \label{equ:sigma_s}
    s_{ij} = \exp\left(\frac{(\beta-1)\|\overline{S}_i^w - \overline{S}_j^w\|_2^2 - \beta\|\overline{S}_i^m - \overline{S}_j^m\|_2^2}{\sigma_s^2}\right),
\end{equation}  
\begin{equation} \label{equ:sigma_l}
    l_{ij} = \exp\left(\frac{-\|\overline{S}_i^p - \overline{S}_j^p\|_2^2}{\sigma_l^2}\right),
\end{equation}  
where \( \beta \) controls the balance between mean and weighted feature contributions, while \( \sigma_s \) and \( \sigma_l \) define the widths of the Gaussian kernels. The resulting weights range between 0 and 1, where a value of 1 indicates maximum similarity. The graph edges are determined using a \( k \)-nearest neighbors (KNN) approach, with edge weights defined as:  
\begin{equation} \label{equ:knn_graph}
    W_{ij} = \begin{cases}
    w_{ij}, & \text{if } i \text{ is one of the } k \text{ nearest neighbors of } j, \\
    & \text{or vice versa}, \\
    0,       & \text{otherwise}.
    \end{cases}
\end{equation}  

During training, a subset of labeled spectral pixels is randomly selected from the original hyperspectral image. The initial label of each superpixel is assigned as the average of the labels of its constituent pixels. If no labeled pixels exist within a superpixel, it remains unassigned initially. The label information is stored in a matrix \( Y \in \mathbb{R}^{K \times c} \), where \( c \) is the number of classes and \( K \) is the total number of superpixels. The entry \( Y_{vl} \) represents the seed label \( l \) for node \( v \). 

The weight matrix and initial labels are then processed using the Local and Global Consistency (LGC) algorithm \citep{zhou2004lgc}, a graph-based semi-supervised learning method that enforces smoothness over the graph structure by minimizing a cost function. The final label matrix \( F \in \mathbb{R}^{K \times c} \) is obtained by minimizing:  
\begin{equation} \label{equ:lcg_loss}
    Q(F) = \frac{1}{2} \sum_{i,j=1}^n W_{ij} \left\|\frac{F_i}{\sqrt{D_{ii}}} - \frac{F_j}{\sqrt{D_{jj}}}\right\|^2 + \frac{\mu}{2} \sum_{i=1}^n \sum_{c=1}^C -y_{ic} \log f_{ic},
\end{equation}  
where \( F^* = \arg\min Q(F) \) represents the optimal label assignment.

\begin{figure*}[h]
    \centering
    \includegraphics[width=\linewidth]{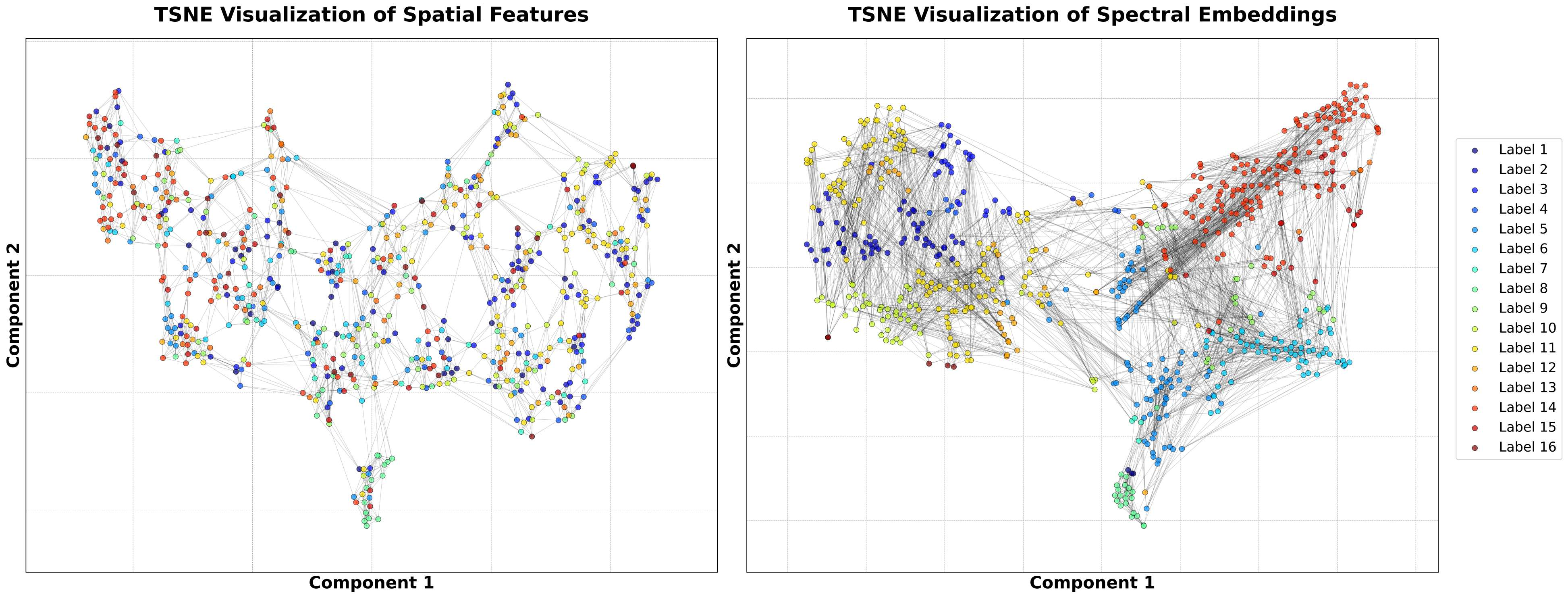}
    \caption{Graph construction visualization for the INDIAN dataset, with node placement based on TSNE embeddings of node features and node labels from GCN inference.}
    \label{fig:indian_graph}
\end{figure*}

\subsection{Multiresolution Graph Learning}

\subsubsection{General Construction}

Multiresolution Graph Networks (MGN), introduced by \citet{hy2019covariant}, offer a framework for analyzing graphs across multiple resolutions. Given an undirected, weighted graph \( \mathcal{G} = (V, E, \mathbf{A}, \mathbf{F}_v) \), where \( V \) and \( E \) denote the sets of nodes and edges, respectively, the graph structure is defined by the adjacency matrix \( \mathbf{A} \in \mathbb{R}^{|V| \times |V|} \). Each node is associated with a feature vector, represented as \( \mathbf{F}_v \in \mathbb{R}^{|V| \times d_v} \), capturing relevant attributes for downstream processing.

\paragraph{Graph Coarsening}
A \emph{K-cluster partition} of a graph divides its nodes into \( K \) mutually exclusive clusters, \( V_1, \ldots, V_K \), where each cluster forms a subgraph. The process of \emph{coarsening} involves constructing a reduced graph \( \widetilde{\mathcal{G}} \), in which each node represents an entire cluster, and edges between these nodes are weighted based on inter-cluster connections in the original graph. This procedure is applied iteratively across multiple levels, resulting in an \emph{L-level coarsening}, where the topmost level condenses the graph into a single node.

\paragraph{Multiresolution Graph Network (MGN)}
MGN operates by iteratively transforming a graph into coarser representations through three key components:
\begin{enumerate}
    \item \textbf{Clustering}: This step partitions the graph into clusters.
    \item \textbf{Encoder}: A graph neural network encodes each cluster into latent node features.
    \item \textbf{Pooling}: Latent features from each cluster are combined into a single vector, which is used to represent the coarser graph at the next level.
\end{enumerate}
These steps are repeated across all levels of resolution, with learnable parameters governing each component. The goal is to predict properties of the original graph by leveraging hierarchical structures.

\subsubsection{Learning to Cluster}

The clustering operation in MGN is differentiable and uses a soft assignment of nodes to clusters, optimized during training via a Gumbel-softmax approximation. This ensures that the clustering procedure can be incorporated into backpropagation for efficient learning.

In summary, MGN provides a scalable way to learn hierarchical representations of graphs by iteratively coarsening them while preserving node and edge information through learnable neural network layers.


\begin{algorithm}
\caption{MultiscaleHSI}\label{alg:multiscalehsi}
\begin{algorithmic}
    \Procedure{MULTISCALEHSI}{Input Image}
        \State Segment superpixels using the Felzenszwalb algorithm
        \State Create mean vector features of the pixels inside superpixels using \ref{equ:mean_filter}
        \State Create weighted vector features using \ref{equ:weighted_filter} and \ref{equ:weight_softmax}
        \State Create centroidal features using \ref{equ:centroid}
        \State Construct K-nearest neighbor graph, where $w_{ij}$ is found by \ref{equ:sigma_s} and \ref{equ:sigma_l}
        \If {$i$ is one of the $k$ nearest neighbors of $j$}
        \State $\mathbf{W}_{ij} = w_{ij}$
        \Else
        \State $\mathbf{W}_{ij} = 0$
        \EndIf
        \State Stratified train-test split (5\% or 10\%) on ground truth pixels
        \If {a superpixel contains a labelled pixel}
        \State Label by their most common training pixel and assign for training
        \Else 
        \State Label randomly as pseudolabel.
        \EndIf
        \State Train MOB-GCN using Local and Global Consistency (LGC), given by \ref{equ:lcg_loss}
    \EndProcedure
\end{algorithmic}
\end{algorithm}

\subsection{Optimal scale selection} \label{sec:optimal}

Selecting the optimal scale for image segmentation is a critical aspect of Object-Based Image Analysis (OBIA) and interpretation. The optimal segmentation scale is defined as the scale at which image objects most accurately correspond to real-world ground features across the entire image. We based on \citet{20-photo-j}’s method for optimal scale selection and applied it for multiple optimal scale selections by selecting ``peaks'' from the relative changes and sort them by value - representing descending impact from number of scales. The optimal scale selection for HrHS image segmentation proceeds as follows:
\begin{enumerate}
    \item Segmenting the HrHS image at different scales and calculating CV;
    \item Detecting and removing outliers using IF algorithm;
    \item Calculating and constructing NN-nCV and NN-nRoC graphs;
    \item Inspecting NN-nRoC graphs and selecting an optimal scale.
\end{enumerate}
The coefficient of variation (CV) for a segment in a single-band image is computed as:  

\[
CV_{p} = \sqrt{\frac{\sum_{i=1}^{N} (x_i - \mu_p)^2}{N}},
\]

where \(CV_{p}\) represents the CV of segment \(p\), \(N\) is the total number of pixels in the segment, \(x_i\) is the intensity value of the \(i\)-th pixel, and \(\mu_p\) is the mean intensity of all pixels within the segment.  

For hyperspectral images, the mean CV across all spectral bands is given by:  

\[
CV_{p} = \frac{\sum_{j=1}^{B} CV_{p,j}}{B},
\]

where \(B\) is the total number of spectral bands, and \(CV_{p,j}\) is the CV of segment \(p\) in band \(j\). The overall average CV across all segments \(P\) in the image is computed as:  

\[
CV_{avg} = \frac{\sum_{p=1}^{P} CV_{p}}{P}.
\]

After computing \(CV_{avg}\), segments with extreme values are filtered out using the Isolation Forest (IF) algorithm \citep{liu2012}, reducing computational cost and memory usage. Once outliers are removed, the NN-nRoC (nearest neighbor-normalized rate of change) is computed using:  

\[
\text{NN-nRoC} = \frac{\left| \left(\frac{\text{NN-nCV}_{n} - \text{NN-nCV}_{n-1}}{\text{NN-nCV}_{n-1}} \right)\right|}{P},
\]

where \(\text{NN-nCV}_{n}\) and \(\text{NN-nCV}_{n-1}\) denote the averaged NN-normalized CV across all segments at scales \(n\) and \(n-1\), respectively.  

Finally, NN-nCV and NN-nRoC graphs are generated for each segmentation method, and the NN-nRoC graph is analyzed to determine the optimal segmentation scale. Peaks in the NN-nRoC graph, where NN-nCV changes significantly, indicate abrupt shifts in intra-segment homogeneity, which correspond to key segmentation scales.

\begin{table}
\centering
\caption{Optimal Scale Selection.}
\label{tab:optimal_scale_selection}
\begin{tabular}{|l|c|}
\hline
\textbf{Dataset} & \textbf{Optimal Scales} \\
\hline
\textbf{INDIAN}   & [42, 24, 17, 8, 4] \\
\hline 
\textbf{SALINAS}  & [55, 31, 23, 14, 10, 4] \\
\hline 
\textbf{PAVIA}    & [71, 17, 14, 8, 5] \\
\hline 
\textbf{BOTSWANA} & [9, 7, 5] \\
\hline 
\textbf{KENNEDY}  & [55, 17, 12, 6] \\
\hline 
\textbf{TORONTO}  & [55, 24, 22, 18] \\
\hline
\end{tabular}
\end{table}

\section{Experiments} \label{sec:experiments}

\subsection{Datasets} \label{sec:datasets}

We use six benchmark HSI datasets to evaluate our approach, which have the following characteristics. The first five can be found at \url{https://www.ehu.eus/ccwintco/index.php/Hyperspectral_Remote_Sensing_Scenes} and UT-HSI-301 can be found at \url{http://vclab.science.uoit.ca/datasets/ut-hsi301/}.

\begin{table*}[h!]
\centering
\caption{Comparison of Hyperspectral Datasets}
\begin{tabular}{|c|c|c|c|c|c|c|}
\hline
\textbf{Data} & \textbf{Location} & \textbf{Spatial resolution (m)} & \textbf{No. of bands} & \textbf{Range (nm)} & \textbf{Labelled data (class)} \\
\hline
Indian Pines Dataset & Indiana, USA & 20 & 200 & 400-2500 & 16  \\
\hline
Salinas & Salinas Valley, California, USA & 3.7 & 200 & 400-2500 & 16  \\
\hline
University of Pavia & Pavia, Italy & 1.3 & 115 & 430-860 & 9  \\
\hline
Botswana & Okavango Delta, Botswana & 30 & 242 & 400-2500 & 14  \\
\hline
Kennedy Space Center & Florida, USA & 18 & 224 & 400-2500 & 13  \\
\hline
University of Toronto & Bolton, Ontario, Canada & 0.3 & 301 & 400-1000 & 4 \\
\hline
\end{tabular}
\label{tab:hyperspectral-comparison}
\end{table*}

\subsection{Evaluation Procotol} \label{sec:evaluation}

Our implementation is done with PyTorch Geometric \citep{pytorch} \citep{pytorchgeometric}, and is available at \url{https://github.com/HySonLab/MultiscaleHSI}. We use the MGN implementation proposed by \cite{hy2019covariant}, with the GCN implementation by \citep{pytorchgeometric}. All experiments were carried out on Google Collaboration, using its T4 GPU with 15 GB of VRAM for hardware acceleration. The environment also featured a dual-core Intel Xeon CPU @ 2.20GHz and 12.72 GB of available system RAM, offering sufficient resources for efficient training and evaluation.

For all experiments, each MOB-GCN model was trained 10 times, and the mean and standard deviation of the results are reported. The optimal number of principal components for the model was selected to retain 99.9\% of the variance in the original image. The performance of each HSI classifier was assessed using three standard evaluation metrics: \textbf{Overall Accuracy (OA)}, \textbf{Average Accuracy (AA)}, and the \textbf{Kappa Coefficient (KA)}.

Our study focuses on answering whether multiscale learning improves learning performance for graph-based hyperspectral classification, so we only validate and compare between a single-scale GCN (with 2 convolution layers, proposed by \citep{kipf2016} and implemented by Pytorch Geometric \cite{pytorchgeometric}), and a multiresolutional graph neural network model, proposed by \citep{Hy_2023}. In addition, we discuss the implications and improvements of our MOB-GCN approach in classifying hyperspectral images.

\subsection{Parameters} \label{sec:parameters}

In our proposed framework, there are seven hyperparameters that come from the four tasks of our framework.
\begin{itemize}
    \item Superpixel construction: $s$.
    \item Feature Extraction: $k$.
    \item Graph construction: $\sigma$, $K$ and $\beta$.
    \item LGC classification: $\mu$, $R$.
\end{itemize}
The fixed parameters are reused from Sellars’ \citep{sellars2020} original implementation.
\begin{table}[h]
\caption{Parameters Description.}
\label{tab:fixed_parameters}
\center
\begin{tabular}{|l|l|r|}
\hline
\textbf{Parameter} & \textbf{Description} & \textbf{Value} \\
\hline
$k$ & Weighted filtering kernel & 15.0 \\
\hline
$\sigma$ & Kernel parameter for constructing $s_{ij}$ & 0.20 \\
\hline
$K$ & k-NN Construction & 8 \\
\hline
$\mu$ & Weighting in the LGC classifier & 0.01 \\
\hline
$\beta$ & Weighting for construction $s_{ij}$ & 0.9 \\
\hline
$s$ & Minimum segmentation siize & See Table \ref{tab:hyperspectral_params} \\
\hline
$R$ & List of resolutions & See Table \ref{tab:hyperspectral_params} \\
\hline
\end{tabular}
\end{table}

For the superpixels construction step, we set the number of superpixels N to be at most 1000 and with a segmentation quality of $\geq 99\%$. Rule of thumb to find a good segmentation size is $width \times height / \textit{number of nodes}$.

\begin{table}[h]
\centering
\caption{Hyperspectral Image Parameters.}
\label{tab:hyperspectral_params}
\begin{tabular}{|c|c|c|c|c|}
\hline
 & \textbf{Shape (h, w)} & \textbf{$s$} & \textbf{\# nodes} & \textbf{$R$} \\
\hline
\textbf{INDIAN} & (145, 145) & 10 & 668 & [16] \\
\hline
\textbf{SALINAS} & (512, 217) & 100 & 239 & [16] \\
\hline
\textbf{PAVIAN} & (1096, 715) & 200 & 921 & [9] \\
\hline
\textbf{BOTSWANA} & (1476, 256) & 200 & 431 & [14] \\
\hline
\textbf{KENNEDY} & (512, 614) & 100 & 522 & [13] \\
\hline
\textbf{TORONTO} & (724, 632) & 200 & 403 & [4] \\
\hline
\end{tabular}
\end{table}

\begin{table}[h!]
\centering
\caption{Number of Parameters across Datasets.}
\label{tab:number_of_params}
\begin{tabular}{|c|c|c|c|}
\hline
\textbf{Dataset} & \textbf{GCN} & \textbf{MGN} & \textbf{MGN (Optimal)} \\ \hline
Indian   & 11,280   & 48,672   & 133,103    \\ \hline
Salinas  & 3,216    & 40,608   & 125,942    \\ \hline
Pavia    & 3,337    & 38,930   & 123,260    \\ \hline
Toronto  & 11,396   & 45,704   & 131,215    \\ \hline
Kennedy  & 13,709   & 50,330   & 140,707    \\ \hline
Botswana & 7,566    & 44,444   & 124,497    \\ \hline
\end{tabular}
\end{table}

\begin{table}[h!]
\centering
\caption{Memory Size across Datasets.}
\label{tab:memory_size}
\begin{tabular}{|c|c|c|c|}
\hline
\textbf{Dataset} & \textbf{GCN (MB)} & \textbf{MGN (MB)} & \textbf{MGN (Optimal, MB)} \\ \hline
Indian   & 0.043  & 0.186  & 0.508  \\ \hline
Salinas  & 0.012  & 0.115  & 0.480  \\ \hline
Pavia    & 0.013  & 0.149  & 0.470  \\ \hline
Toronto  & 0.043  & 0.174  & 0.501  \\ \hline
Kennedy  & 0.052  & 0.192  & 0.537  \\ \hline
Botswana & 0.029  & 0.170  & 0.475  \\ \hline
\end{tabular}
\end{table}

We conducted experiments using two methods for selecting learned resolutions: one based on the number of classes in the dataset and the other on identifying the optimal resolutions. A more detailed discussion on determining these optimal resolutions will be provided in the Discussion section.

\subsection{Results} \label{sec:results}

Our experiments are divided into two parts. First, we evaluate the classification accuracy of our proposed framework against the baseline classifiers mentioned earlier. Given the semi-supervised nature of our approach, we assess classification performance using limited training data, specifically 5\%, 10\%, and 20\% of the sample data. Second, we analyze visual classification maps to interpret and compare the performance of our multiresolution model with single-scale GCNs.

\begin{table}
\caption{OA (\%) AA (\%) and Kappa (\%) of ten consecutive experiments with 5\% sample data (* means optimal scales selected.)}
\centering
\small
\begin{tabular}{|lccc|}
\hline
\multicolumn{4}{|c|}{\textbf{INDIAN}} \\
\hline
\textbf{Model} & \textbf{OA} & \textbf{AA} & \textbf{Kappa} \\
\hline
GCN & 92.85 $\pm$ 0.04 & 86.25 $\pm$ 0.02 & 91.82 $\pm$ 0.00 \\
MOB-GCN & \textbf{94.39 $\pm$ 0.02} & \textbf{94.66 $\pm$ 0.03} & \textbf{93.56 $\pm$ 0.00} \\
MOB-GCN (*) & 94.28 $\pm$ 0.00 & 94.53 $\pm$ 0.00 & 93.49 $\pm$ 0.00 \\
\hline
\multicolumn{4}{|c|}{\textbf{SALINAS}} \\
\hline
\textbf{Model} & \textbf{OA} & \textbf{AA} & \textbf{Kappa} \\
\hline
GCN & 89.35 $\pm$ 0.02 & 91.35 $\pm$ 0.02 & 88.09 $\pm$ 0.02 \\
MOB-GCN & 98.52 $\pm$ 0.00 & \textbf{99.01 $\pm$ 0.00} & 98.35 $\pm$ 0.00 \\
MOB-GCN (*) & \textbf{98.85 $\pm$ 0.00} & 99.11 $\pm$ 0.00 & \textbf{98.72 $\pm$ 0.00} \\
\hline
\multicolumn{4}{|c|}{\textbf{PAVIA}} \\
\hline
\textbf{Model} & \textbf{OA} & \textbf{AA} & \textbf{Kappa} \\
\hline
GCN & 95.63 $\pm$ 0.00 & 85.03 $\pm$ 0.01 & 93.79 $\pm$ 0.00 \\
MOB-GCN & 96.72 $\pm$ 0.00 & 90.52 $\pm$ 0.00 & 95.35 $\pm$ 0.00 \\
MOB-GCN (*) & \textbf{96.79 $\pm$ 0.00} & \textbf{90.76 $\pm$ 0.00} & \textbf{95.45 $\pm$ 0.00} \\
\hline
\multicolumn{4}{|c|}{\textbf{KENNEDY}} \\
\hline
\textbf{Model} & \textbf{OA} & \textbf{AA} & \textbf{Kappa} \\
\hline
GCN & 80.10 $\pm$ 0.01 & 68.93 $\pm$ 0.01 & 78.32 $\pm$ 0.01 \\
MOB-GCN & 92.84 $\pm$ 0.01 & 87.10 $\pm$ 0.02 & 92.02 $\pm$ 0.01 \\
MOB-GCN (*) & \textbf{93.69 $\pm$ 0.01} & \textbf{89.70 $\pm$ 0.01} & \textbf{92.97 $\pm$ 0.01} \\
\hline
\multicolumn{4}{|c|}{\textbf{BOTSWANA}} \\
\hline
\textbf{Model} & \textbf{OA} & \textbf{AA} & \textbf{Kappa} \\
\hline
GCN & 91.96 $\pm$ 0.02 & 92.08 $\pm$ 0.02 & 91.23 $\pm$ 0.02 \\
MOB-GCN & 93.34 $\pm$ 0.00 & 93.44 $\pm$ 0.01 & 93.00 $\pm$ 0.00 \\
MOB-GCN (*) & \textbf{93.40 $\pm$ 0.00} & \textbf{93.17 $\pm$ 0.00} & \textbf{92.85 $\pm$ 0.00} \\
\hline
\multicolumn{4}{|c|}{\textbf{TORONTO}} \\
\hline
\textbf{Model} & \textbf{OA} & \textbf{AA} & \textbf{Kappa} \\
\hline
GCN & 96.45 $\pm$ 0.93 & 96.86 $\pm$ 1.01 & 95.08 $\pm$ 1.29 \\
MOB-GCN & 97.41 $\pm$ 0.01 & 97.88 $\pm$ 0.02 & 96.42 $\pm$ 0.02 \\
MOB-GCN (*) & \textbf{97.42 $\pm$ 0.01} & \textbf{97.88 $\pm$ 0.02} & \textbf{96.42 $\pm$ 0.02} \\
\hline
\end{tabular}
\label{table:5_percent_results}
\end{table}

\begin{figure*}[h]
    \centering
    \begin{minipage}{0.48\linewidth}  
        \centering
        \subfloat[INDIAN]{%
            \includegraphics[width=\linewidth]{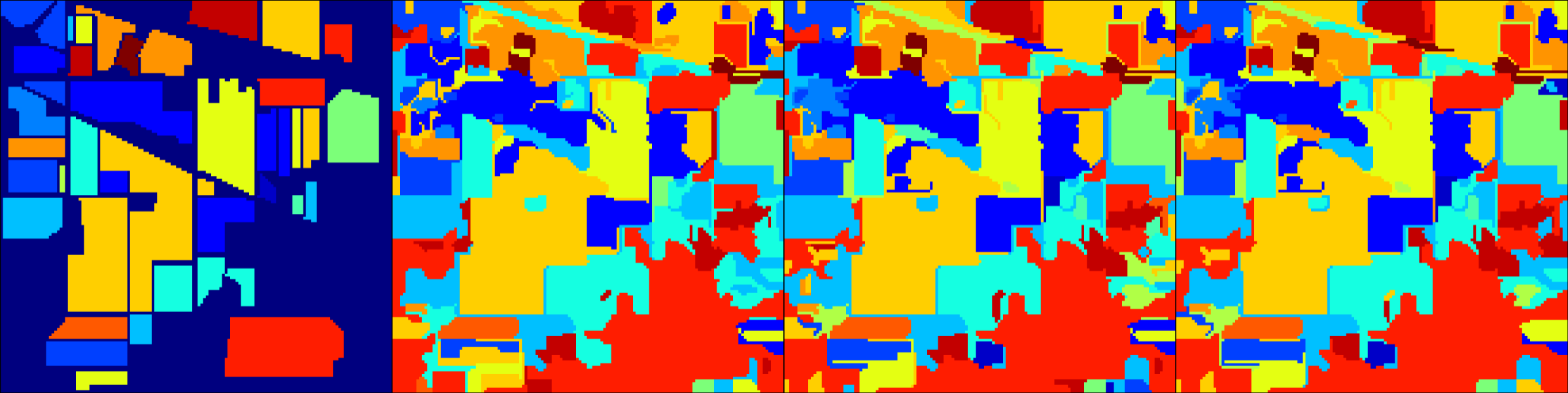}
        }
        \vspace{0.3em}

        \subfloat[SALINAS]{%
            \includegraphics[width=\linewidth]{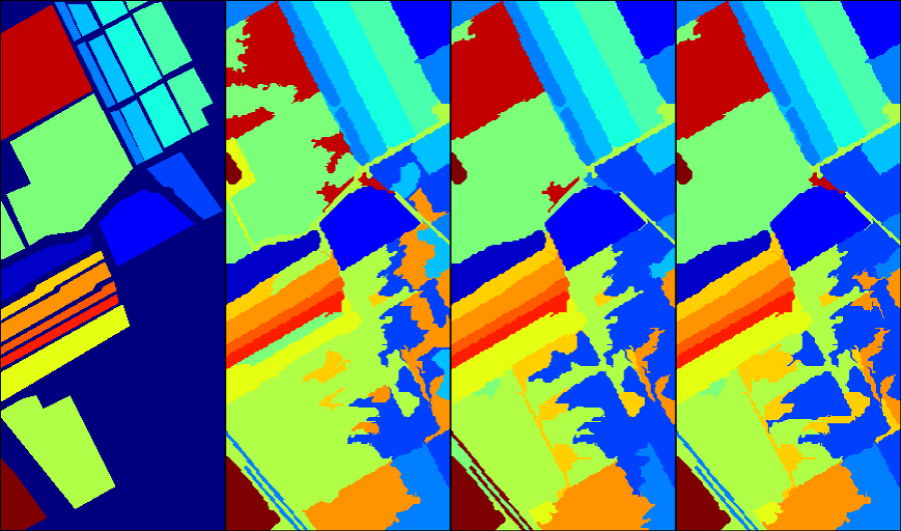}
        }
        \vspace{0.3em}

        \subfloat[PAVIA]{%
            \includegraphics[width=\linewidth]{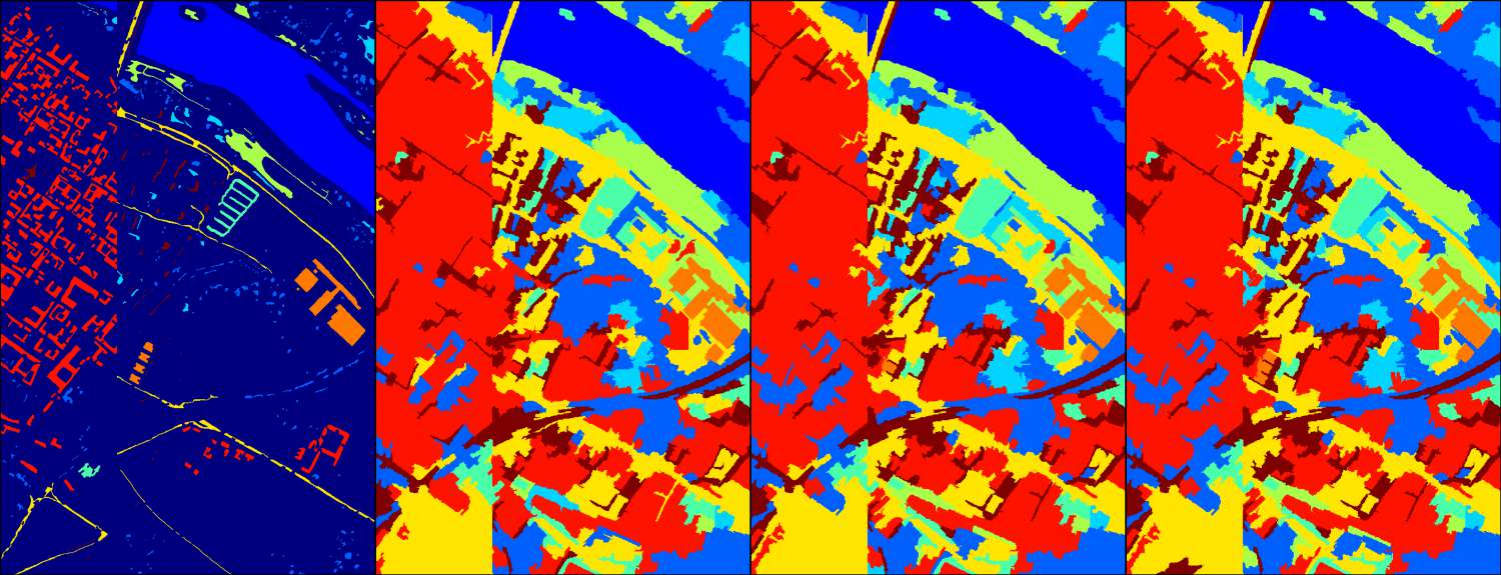}
        }
        \vspace{0.3em}

        \subfloat[KENNEDY]{%
            \includegraphics[width=\linewidth]{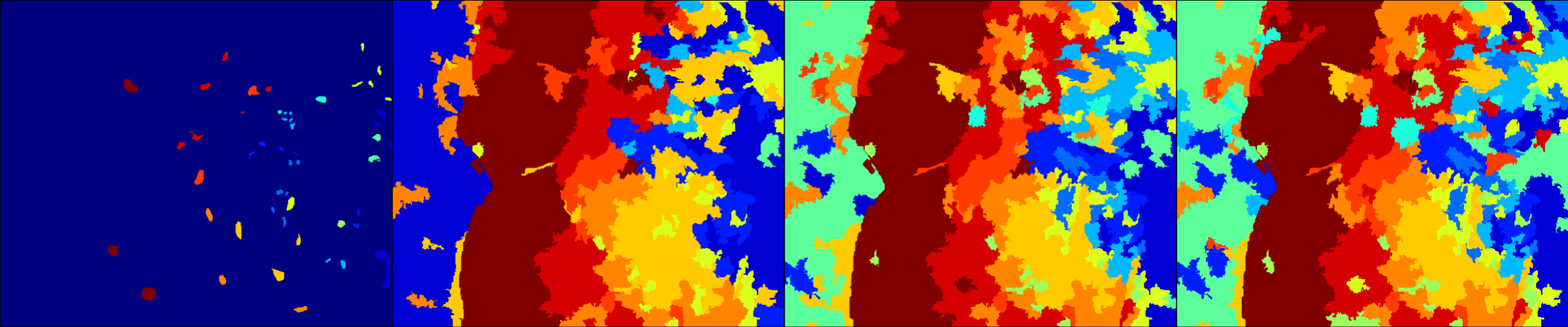}
        }
    \end{minipage}
    \hfill
    \begin{minipage}{0.48\linewidth}  
        \centering
        \subfloat[BOTSWANA]{%
            \includegraphics[width=\linewidth]{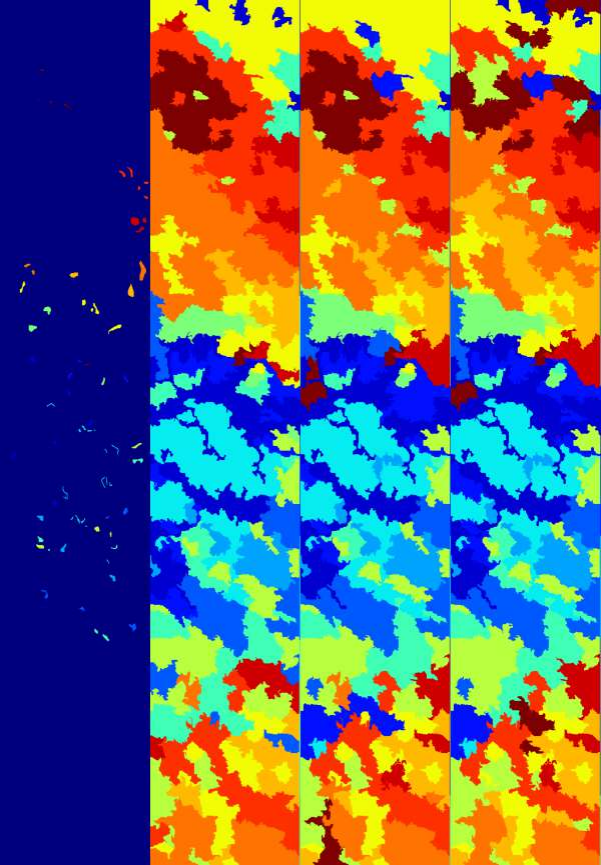}
        }
        \vspace{0.3em}

        \subfloat[TORONTO]{%
            \includegraphics[width=\linewidth]{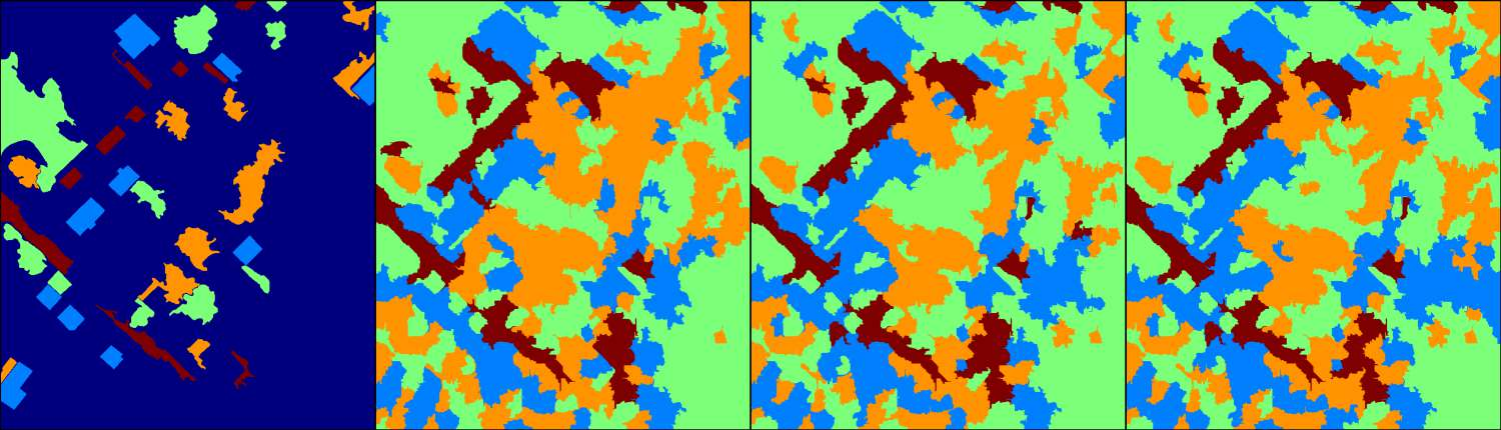}
        }
    \end{minipage}

    \caption{Comparison of classification maps for different datasets on 5\% sample data. Each subfigure shows ground truth labels and classification maps from GCN, MOB-GCN, unoptimized, and optimized models.}
    \label{fig:collage}
\end{figure*}
\section{Discussions} \label{sec:discussions}

\subsection{Classification Results of MOB-GCN} 

The MOB-GCN model, especially in its optimized form (MOB-GCN (Optimal)), consistently achieved the highest OA compared to single-scale GCNs and the non-optimized MOB-GCN. This indicates that incorporating multiscale information and selecting optimal scales significantly enhances classification performance.

See table \ref{table:5_percent_results}, with only 5\% of the data used for training , the MOB-GCN (Optimal) achieved an OA of 94.28\% on the Indian Pines dataset, while the single-scale GCN only achieved 92.85\%. For the Salinas dataset with 5\% training data, the MOB-GCN (Optimal) showed a substantial improvement with an OA of 98.85\%, compared to the GCN's 89.35\%. Similar trends were observed across other datasets like Pavia, Kennedy Space Center, Botswana and University of Toronto, with the optimized MOB-GCN consistently outperforming the other models.

The superior performance of the MOB-GCN (Optimal) is not limited to small training datasets, with the optimized model performing the best across all sample sizes (5\%, 10\%, and 20\%). This shows the model's robustness and adaptability to varying amounts of training data. The MOB-GCN models show marked improvement over the GCN, especially with smaller sample sizes.

The key to the MOB-GCN's performance is its ability to integrate features extracted from multiple segmentation scales, enabling it to capture both fine-grained details and broader contextual information. The multiscale approach allows for a more comprehensive understanding of complex structures within hyperspectral images, which leads to more accurate classification outcomes. By using superpixels as nodes in the graph, the MOB-GCN reduces the computational overhead associated with processing large hyperspectral images\cite{sellars2020}. 

\subsection{Comparing MOB-GCN with Single-Scale Methods}

The MOB-GCN consistently surpassed the single-scale GCN in performance across all datasets. The single-scale GCN often exhibited lower classification accuracy and a higher prevalence of speckle noise in the output maps, highlighting its limitations in capturing the hierarchical
spatial-spectral relationships within HSI data. In contrast, the MOB-GCN, which integrates information from multiple scales, generated smoother and more accurate classification maps.

\subsection{Impact of scale on MOB-GCN Performance}

The automatic optimal scale selection method, which identifies the most informative segmentation scales, is crucial to the success of the MOB-GCN. This method determines optimal scales based on the relative changes in the coefficient of variation (CV) across different segmentation scales. It selects the "peaks" of these changes, which indicate significant variations in image object heterogeneity. Our experiments demonstrated that combining features from 4–6 optimal segmentation scales was generally sufficient to achieve the desired classification performance across most datasets. The specific optimal scales for each dataset are provided in Table VI. For instance, the optimal scales for the Indian Pines dataset were 42, 24, 17, 8, and 4, while for the Salinas dataset, they were 55, 31, 23, 14, 10, and 4.

\begin{figure}[h!]
    \centering
    \includegraphics[width=\linewidth]{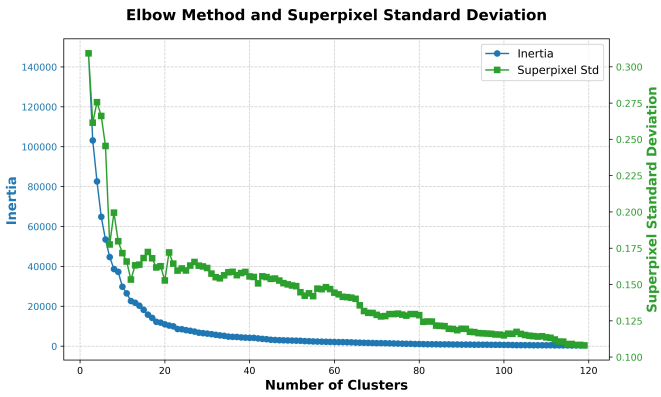}
    \caption{Inertia and Superpixel standard deviation for K-means clustering assessment on the SALINAS dataset.}
    \label{fig:inertia_standard_deviation}
\end{figure}

\begin{figure}[h!]
    \centering
    \includegraphics[width=\linewidth]{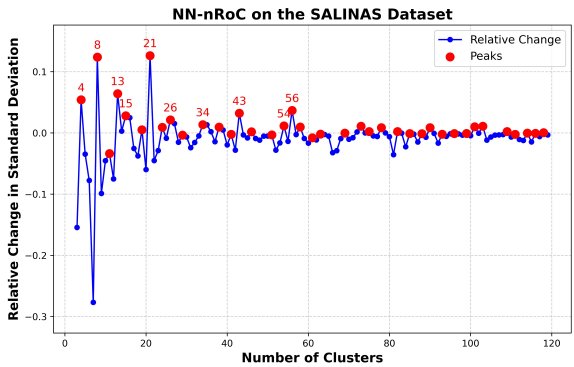}
    \caption{NN-nRoC on every number of clusters on the SALINAS Dataset.
}
    \label{fig:nn_nRoC}
\end{figure}

\begin{figure}[h!]
    \centering
    \includegraphics[width=\linewidth]{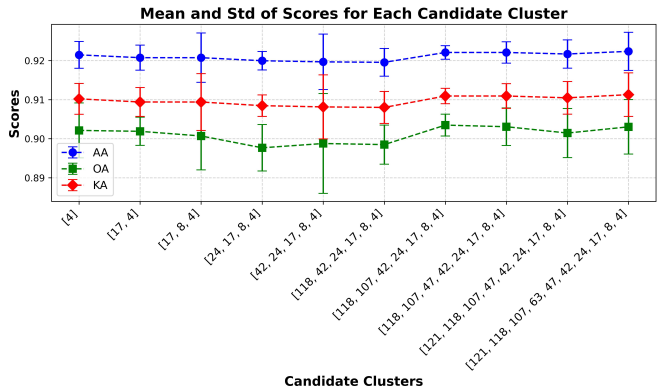}
    \caption{MGN Performance on training with 10 sizes of candidate clusters, each are chosen by descending NN-nROC value. Each test is repeated 5 times, and the last 3 are omitted due to poor performance. The test was performed on the SALINAS dataset.}
    \label{fig:candidate_size}
\end{figure}

\section{Conclusion} \label{sec:conclusion}

This novel MOB-GCN leverages a multiresolution approach, inspired by Multiresolution Graph Networks (MGN), to capture both fine-grained details and global context in HSI data. By integrating features from multiple segmentation scales, MOB-GCN achieves higher classification accuracy than single-scale GCN models. The MOB-GCN (Optimal) model, which incorporates an automatic optimal scale selection, consistently achieves the highest Overall Accuracy (OA) across various datasets and sample sizes. For instance, with only 5\% training data, the optimized MOB-GCN achieved significantly higher OA on the Salinas (98.85\% vs. 89.35\%) and Kennedy (93.69\% vs. 80.10\%) datasets, compared to the single-scale GCN model.

MOB-GCN is particularly advantageous when labeled data is limited, a common challenge in remote sensing image classification due to the time-consuming and labor-intensive nature of field data collection. By leveraging information from multiple resolutions, MOB-GCN enhances robustness and adaptability, achieving superior classification performance even with scarce training data.

MOB-GCN is optimized for computational efficiency by representing superpixels as graph nodes, significantly reducing the number of nodes compared to pixel-based methods. This smaller graph size allows for faster matrix inversions and other computations. Additionally, the multiresolution process, which includes graph coarsening at higher levels, further accelerates processing, enhancing the scalability of the approach for HSI analysis.

The automatic optimal scale selection method, based on analyzing the coefficient of variation (CV) across different segmentation scales, is essential to MOB-GCN's superior performance. By identifying the most significant scales, the model effectively captures important spatial-spectral features. Experiments show that combining features from 4–6 optimal segmentation scales was generally sufficient to achieve the desired classification performance across most datasets.

It is important to highlight that the benefits of multiresolution networks like MOB-GCN tend to diminish as the size and complexity of hyperspectral images increase. For very large HSI datasets, conventional GNNs may provide a more efficient alternative, as they can handle large graphs without the overhead of constructing multiscale hierarchies.  

Overall, MOB-GCN presents a practical approach for hyperspectral image analysis, particularly in applications that demand both high accuracy and computational efficiency. Its strong performance with limited training data further enhances its applicability, addressing the challenges associated with acquiring labeled hyperspectral data in remote sensing.


\bibliography{paper}


\appendices
\onecolumn

\section{INDIAN at 5\% Sample Data} \label{sec:apx_indian}

\begin{figure}[h!]
\centering
\begin{tabular}{cc}
\includegraphics[width=0.5\textwidth]{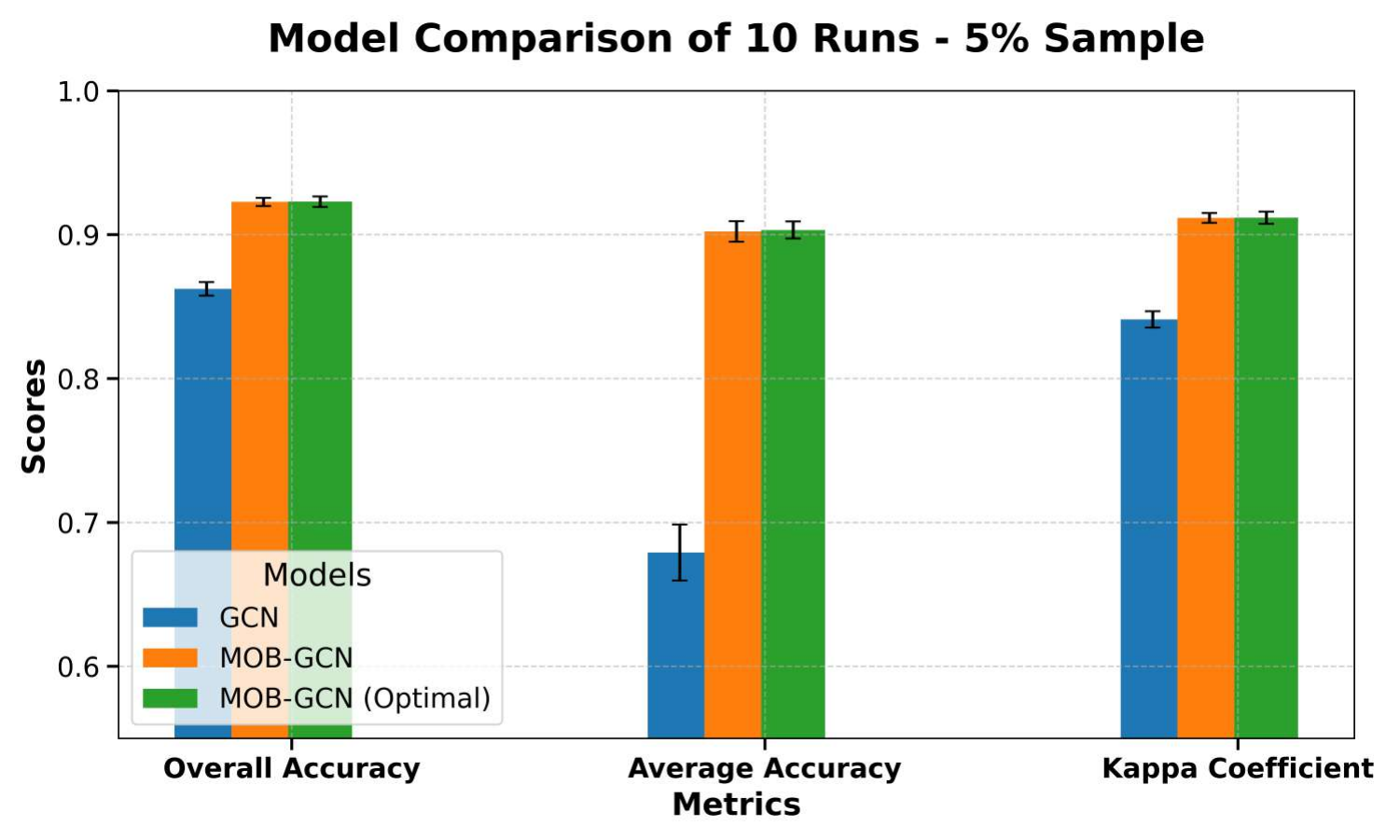} &
\includegraphics[width=0.5\textwidth]{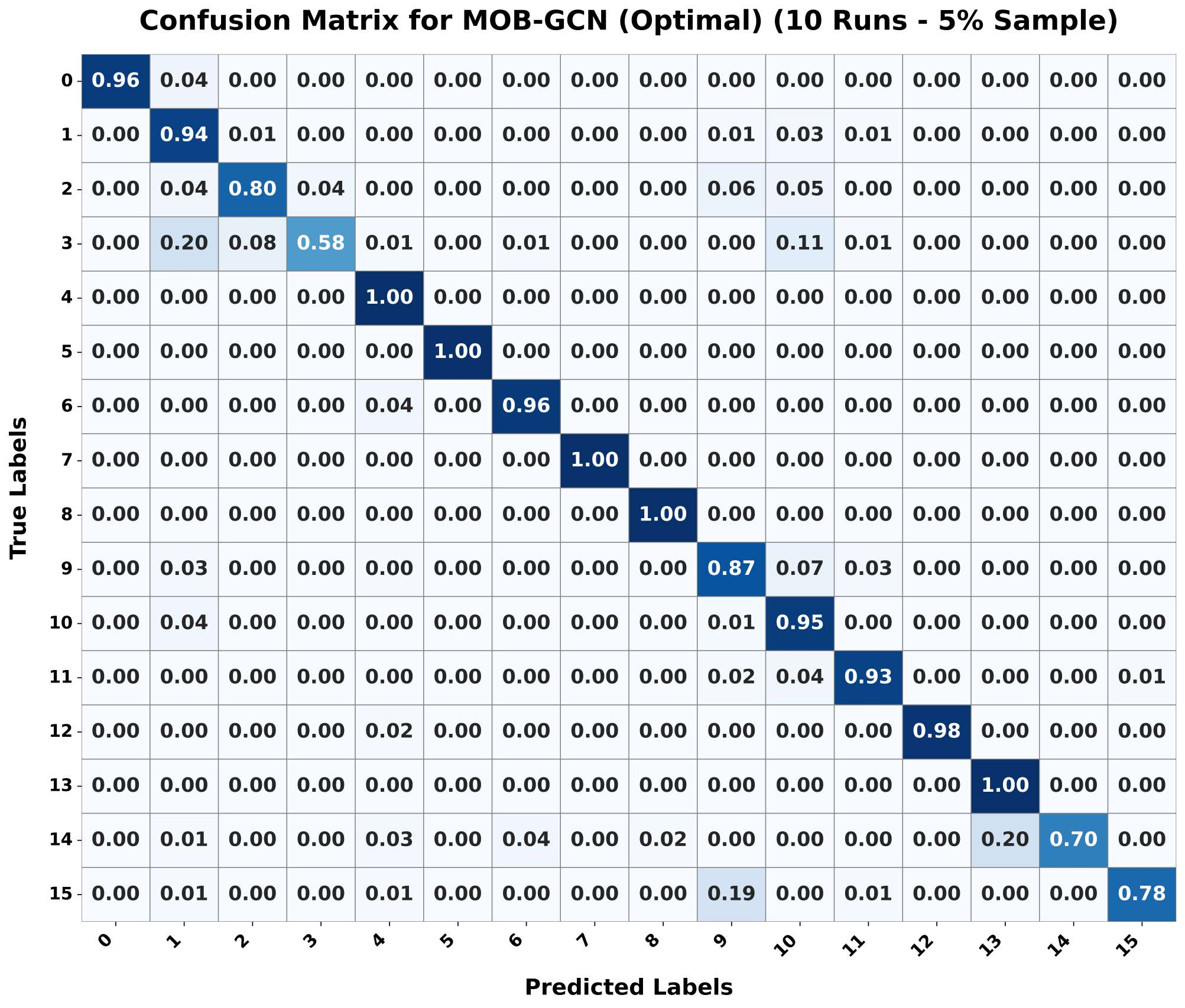} \\
\end{tabular}
\caption{INDIAN on 5\% Sample Data. Model comparison and Average Confusion Matrix from MOB-GCN (Optimal).}
\label{fig:indian_metrics}
\end{figure}

\begin{figure}[h!]
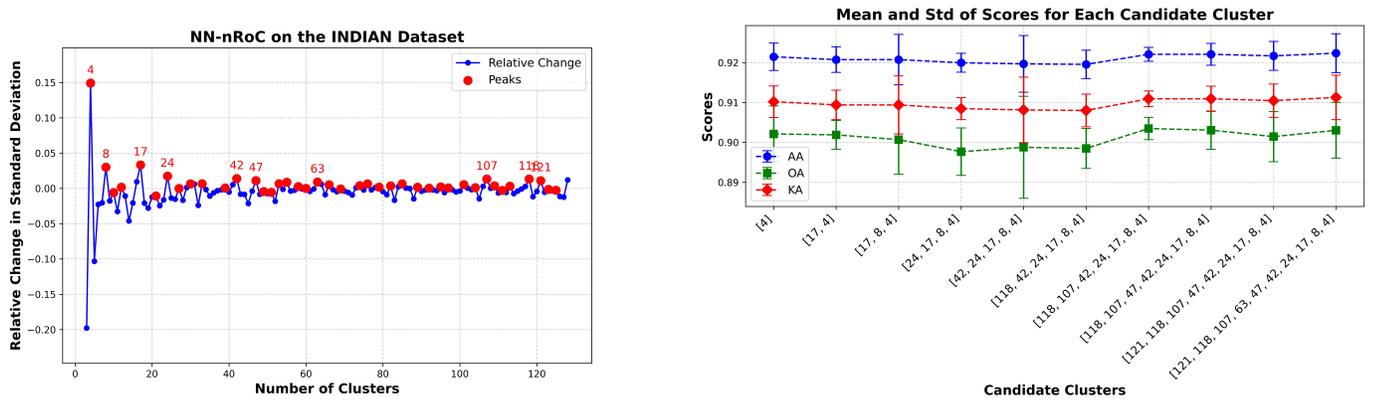

\centering
\begin{tabular}{cc}
\includegraphics[width=0.5\textwidth]{appendix/images/INDIAN/NN-nROC.pdf} &
\includegraphics[width=0.5\textwidth]{appendix/images/INDIAN/candidate_clusters.pdf} \\
\end{tabular}
\caption{INDIAN on 5\% Sample Data. NN-nRoC on every number of clusters on the INDIAN Dataset. Peaks on INDIAN’ NN-nROC are placed at 4, 17, 8, 24, 42, 84, 102, 47, 75, 55 with descending value.}
\label{fig:indian_scales}
\end{figure}

\section{SALINAS at 5\% Sample Data} \label{sec:apx_salinas}

\begin{figure}[h!]
\centering
\begin{tabular}{cc}
\includegraphics[width=0.5\textwidth]{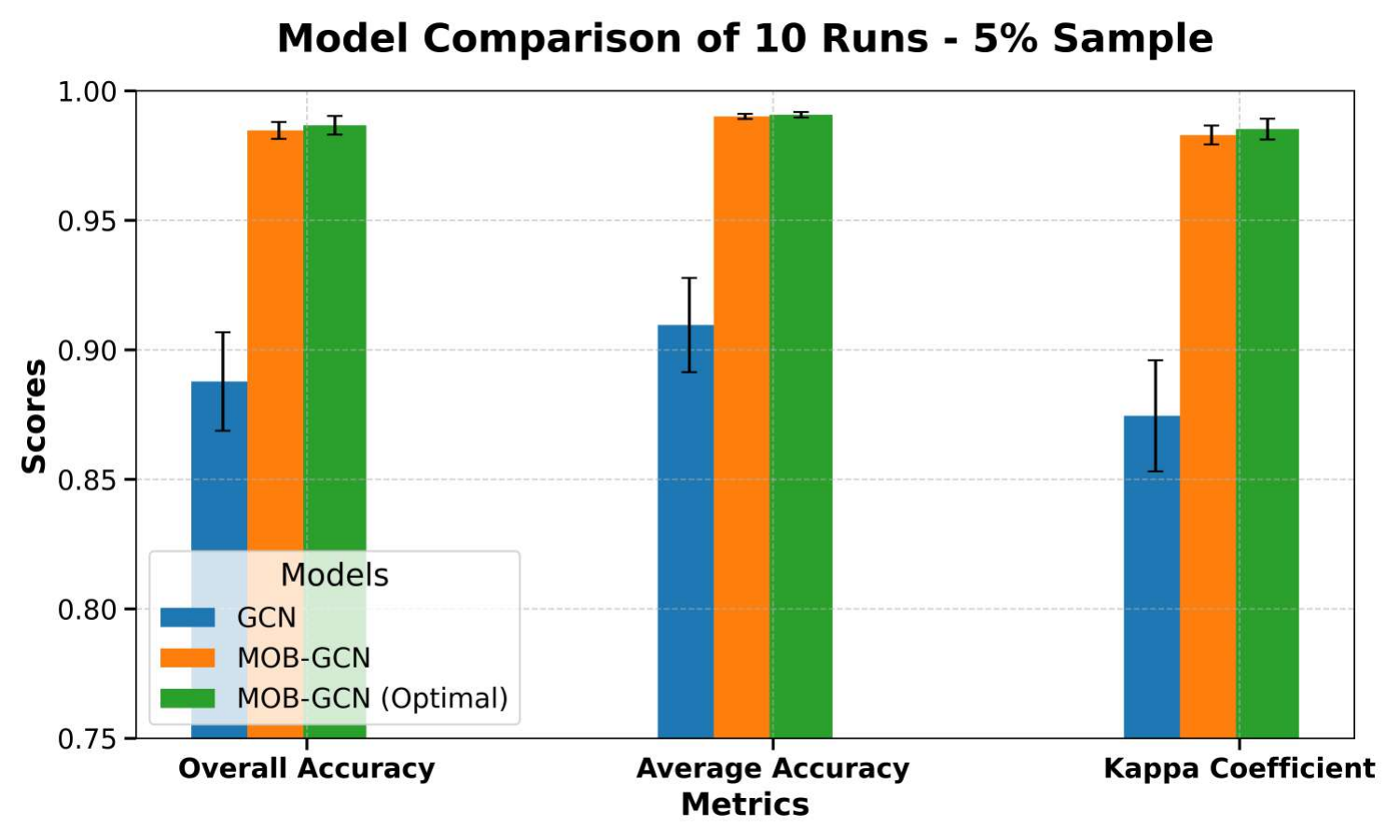} &
\includegraphics[width=0.5\textwidth]{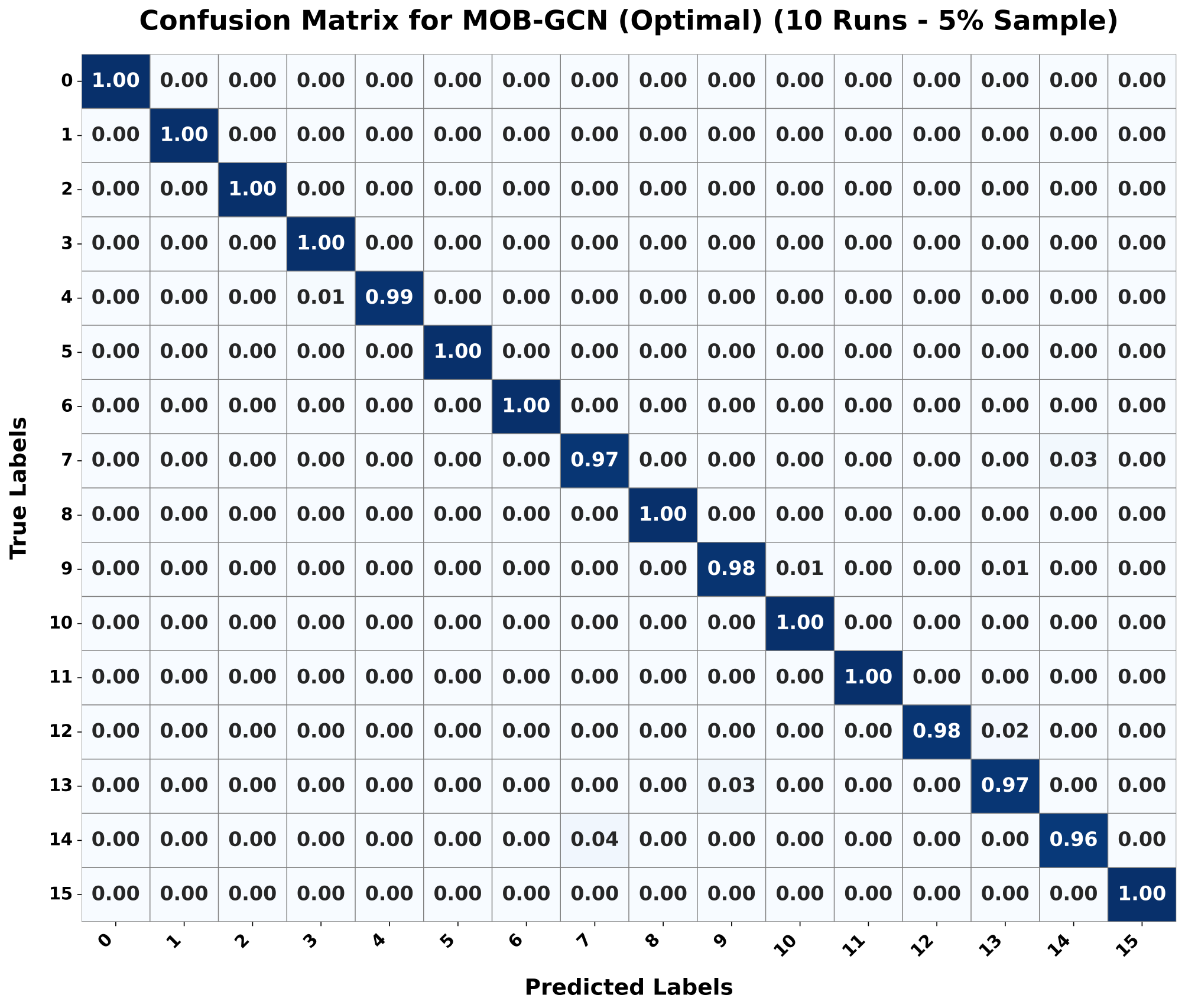} \\
\end{tabular}
\caption{SALINAS on 5\% Sample Data. Model comparison and Average Confusion Matrix from MOB-GCN (Optimal).}
\label{fig:salinas_metrics}
\end{figure}

\begin{figure}[h!]
\centering
\begin{tabular}{cc}
\includegraphics[width=0.5\textwidth]{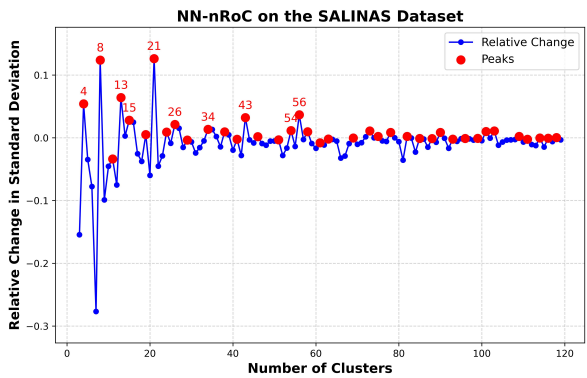} &
\includegraphics[width=0.5\textwidth]{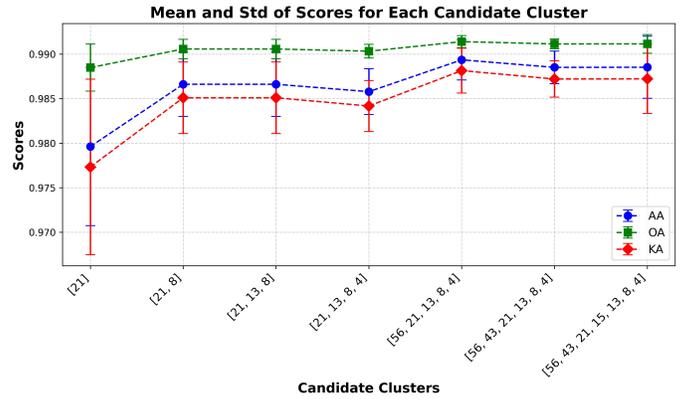} \\
\end{tabular}
\caption{SALINAS on 5\% Sample Data. NN-nRoC on every number of clusters on the INDIAN Dataset. Peaks on SALINAS’ NN-nROC are placed at 4, 55, 10, 23, 14, 31, 88, 33, 78, 17, 40 with descending value.}
\label{fig:salinas_scales}
\end{figure}

\section{PAVIA at 5\% Sample Data} \label{sec:apx_pavia}

\begin{figure}[h!]
\centering
\begin{tabular}{cc}
\includegraphics[width=0.5\textwidth]{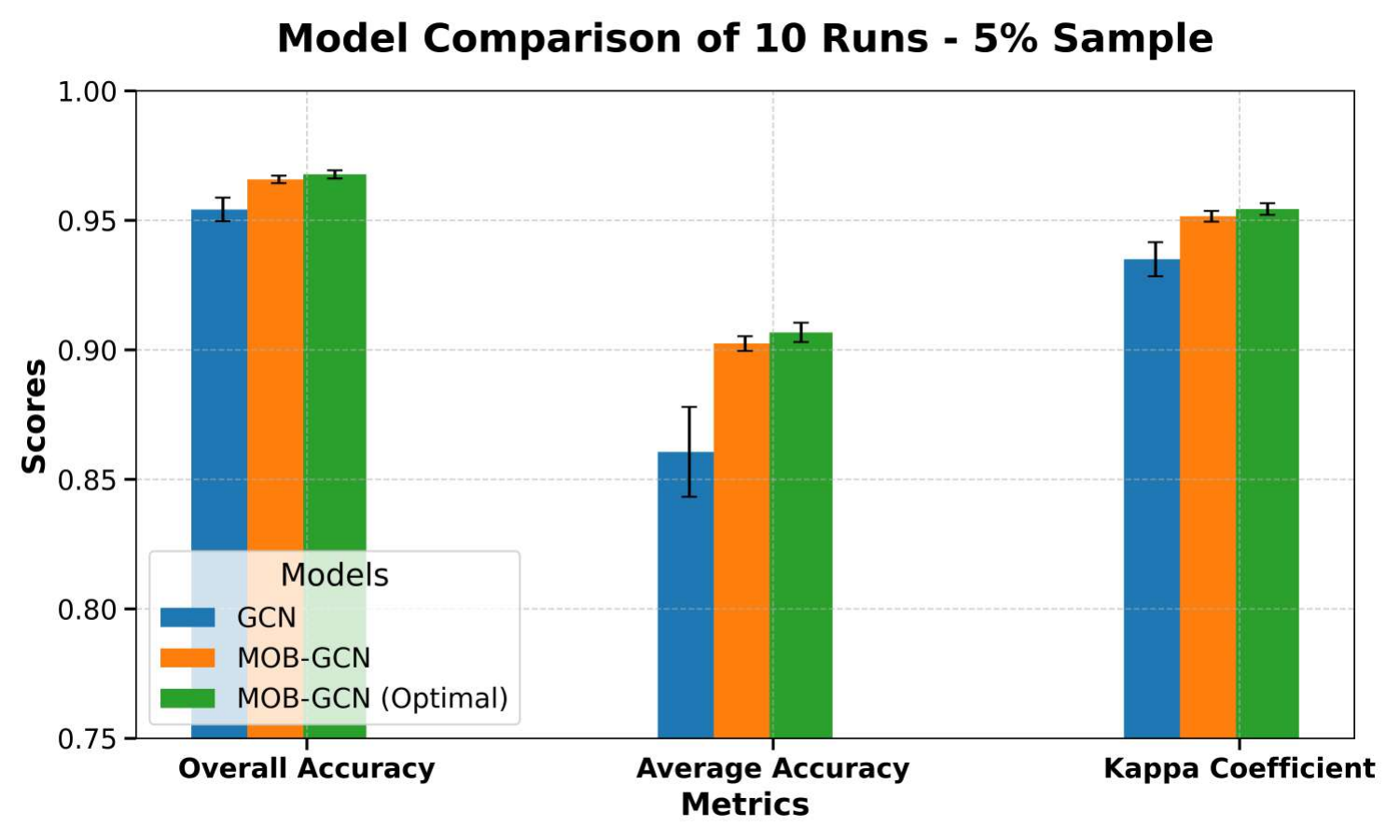} &
\includegraphics[width=0.5\textwidth]{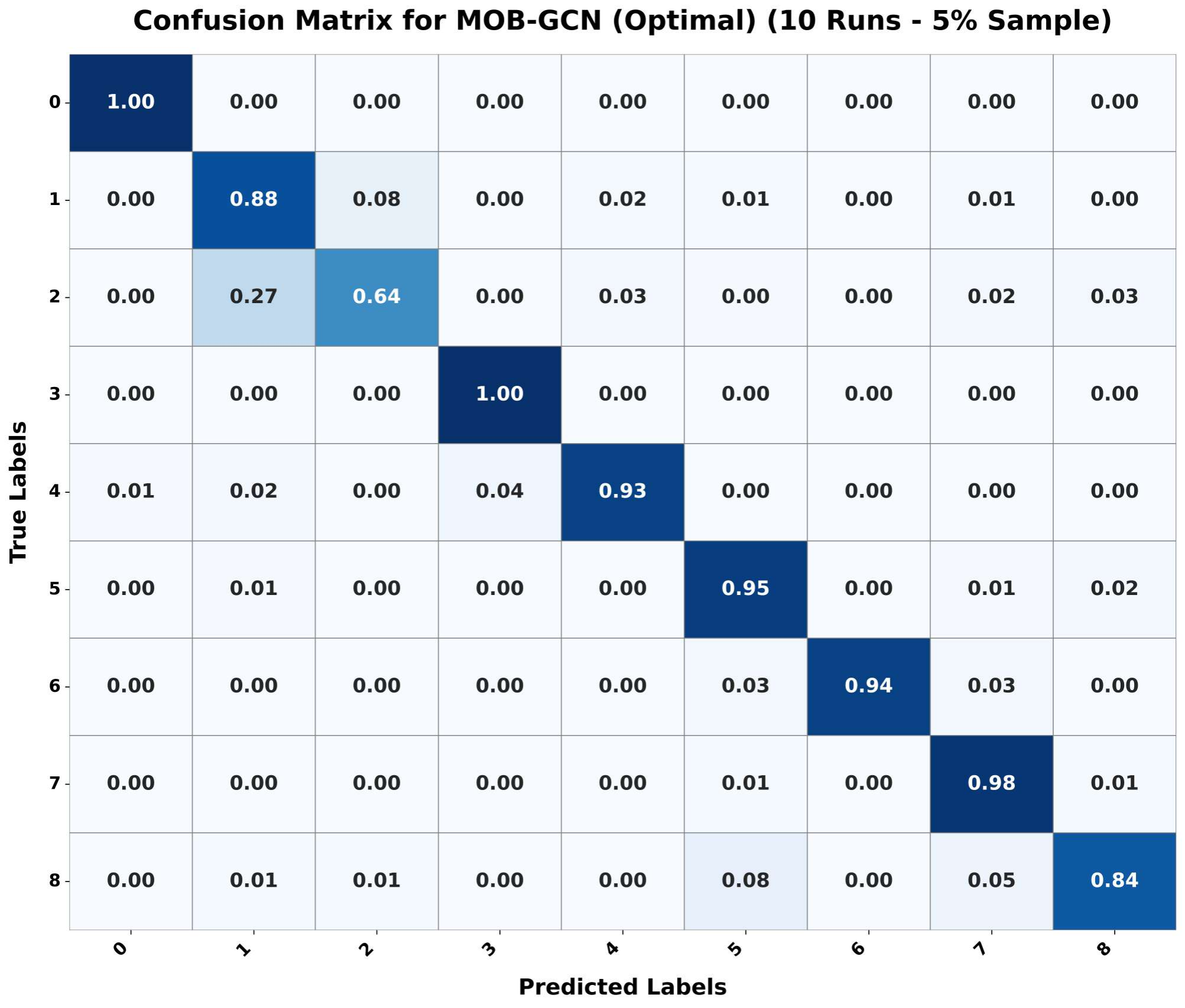} \\
\end{tabular}
\caption{PAVIA on 5\% Sample Data. Model comparison and Average Confusion Matrix from MOB-GCN (Optimal).}
\label{fig:pavia_metrics}
\end{figure}

\begin{figure}[h!]
\centering
\begin{tabular}{cc}
\includegraphics[width=0.5\textwidth]{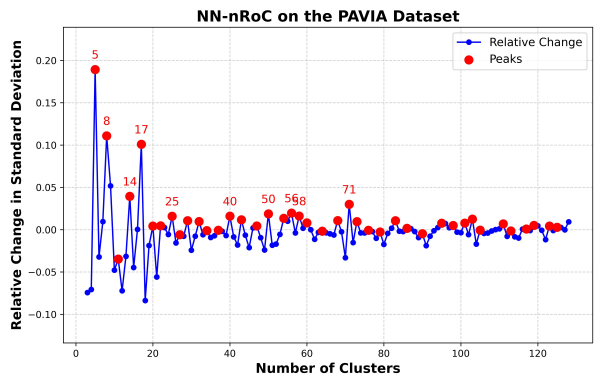} &
\includegraphics[width=0.5\textwidth]{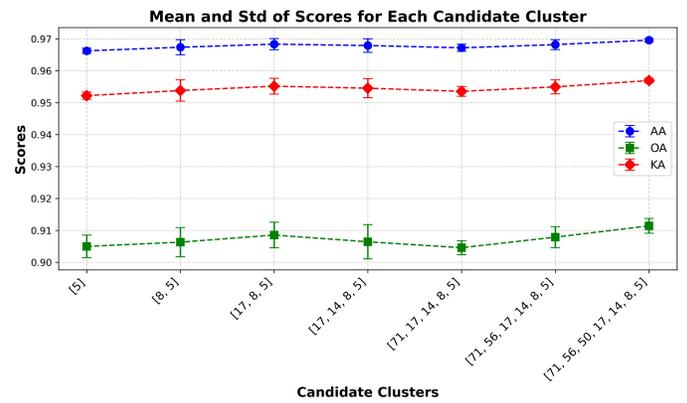} \\
\end{tabular}
\caption{PAVIA on 5\% Sample Data. NN-nRoC on every number of clusters on the PAVIA Dataset. Peaks on PAVIA’ NN-nROC are placed at 5, 8, 17, 14, 71, 56, 50, 58, 40, 25 with descending value.}
\label{fig:pavia_scales}
\end{figure}
\section{KENNEDY at 5\% Sample Data} \label{sec:apx_kennedy}

\begin{figure}[h!]
\centering
\begin{tabular}{cc}
\includegraphics[width=0.5\textwidth]{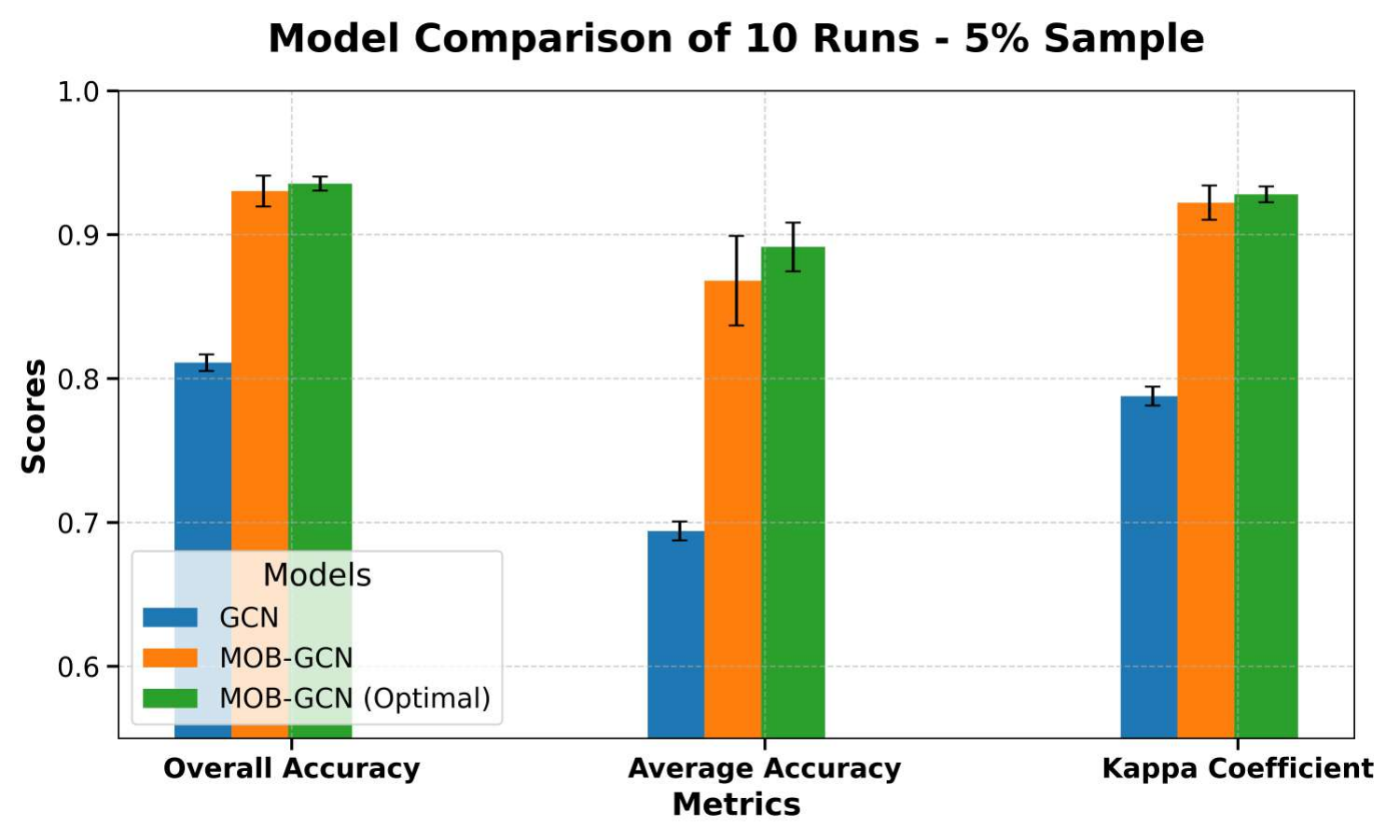} &
\includegraphics[width=0.5\textwidth]{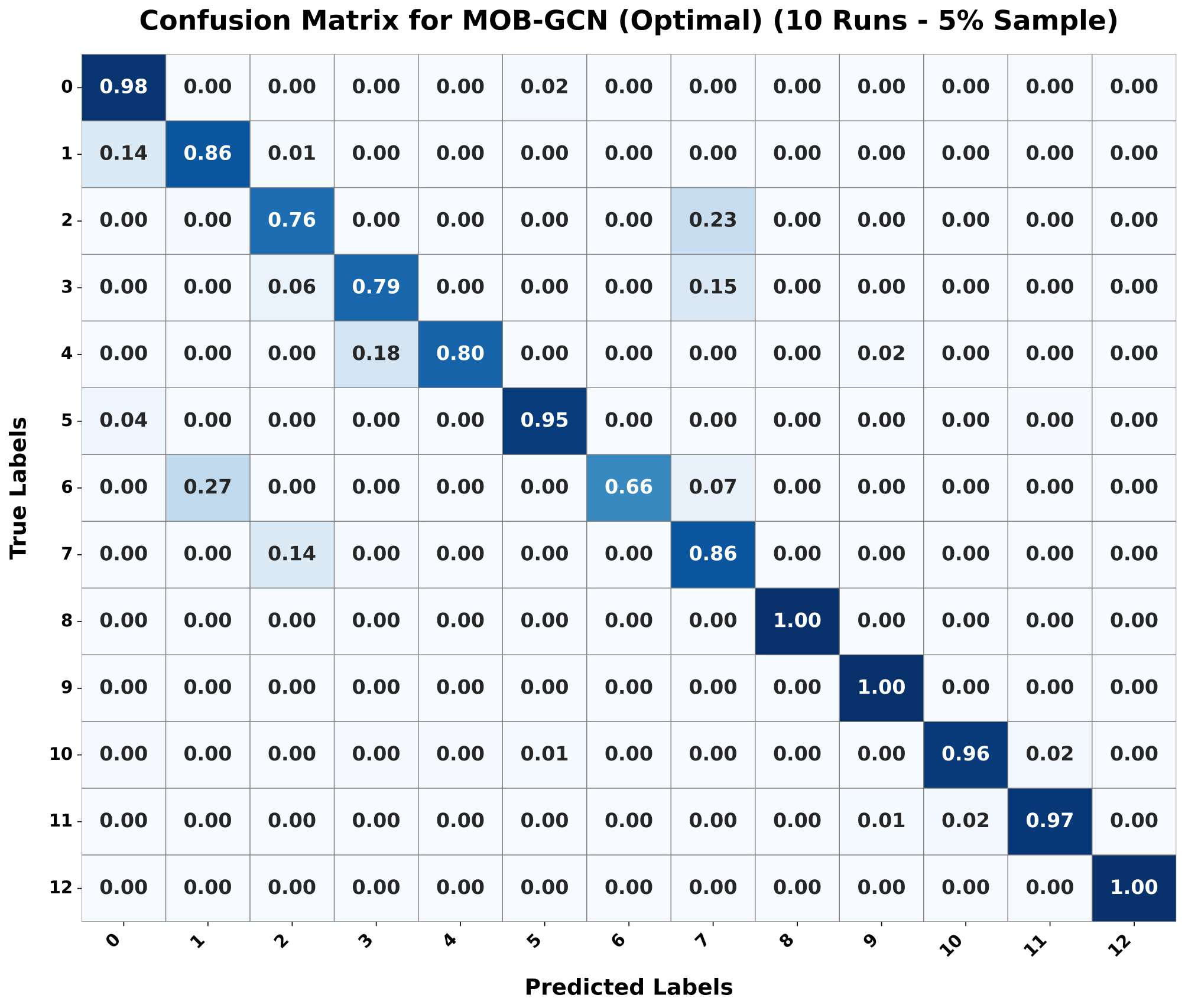} \\
\end{tabular}
\caption{KENNEDY on 5\% Sample Data. Model comparison and Average Confusion Matrix from MOB-GCN (Optimal).}
\label{fig:kennedy_metrics}
\end{figure}

\begin{figure}[h!]
\centering
\begin{tabular}{cc}
\includegraphics[width=0.5\textwidth]{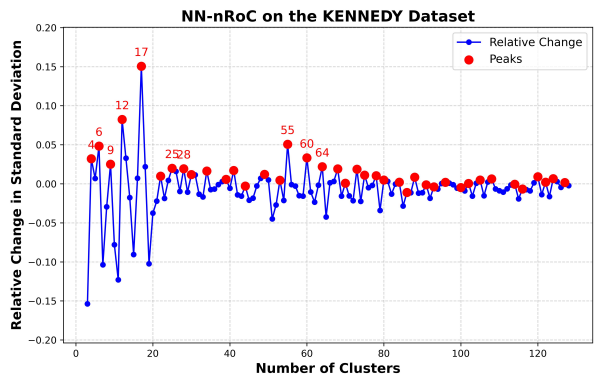} &
\includegraphics[width=0.5\textwidth]{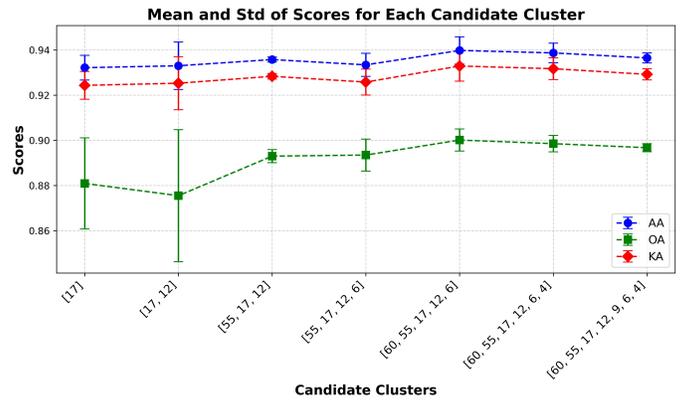} \\
\end{tabular}
\caption{KENNEDY on 5\% Sample Data. NN-nRoC on every number of clusters on the KENNEDY Dataset. Peaks on PAVIA’ NN-nROC are placed at 17, 12, 55, 6, 60, 4, 9, 64, 25, 28 with descending value.}
\label{fig:kennedy_scales}
\end{figure}
\section{BOTSWANA at 5\% Sample Data} \label{sec:apx_botswana}

\begin{figure}[h!]
\centering
\begin{tabular}{cc}
\includegraphics[width=0.5\textwidth]{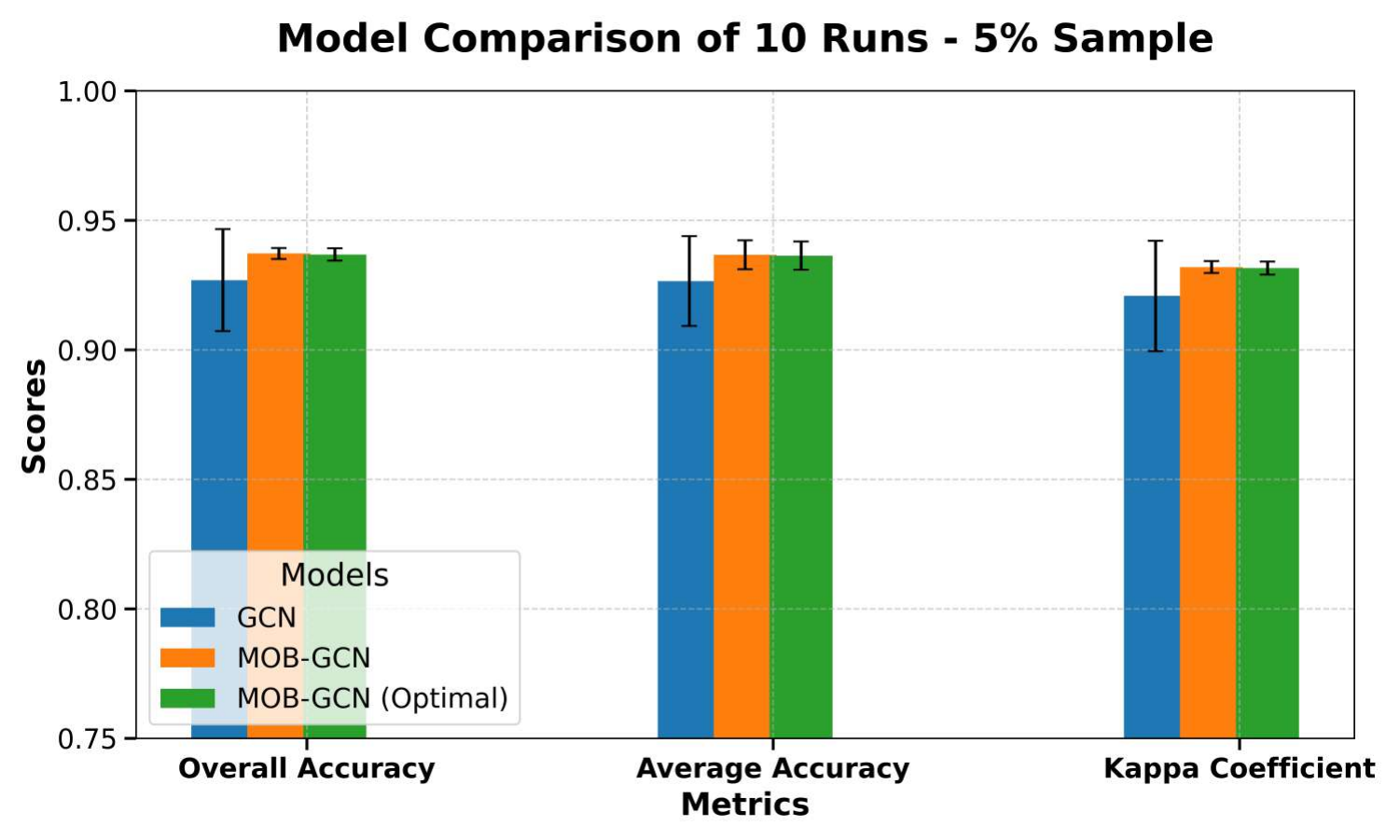} &
\includegraphics[width=0.5\textwidth]{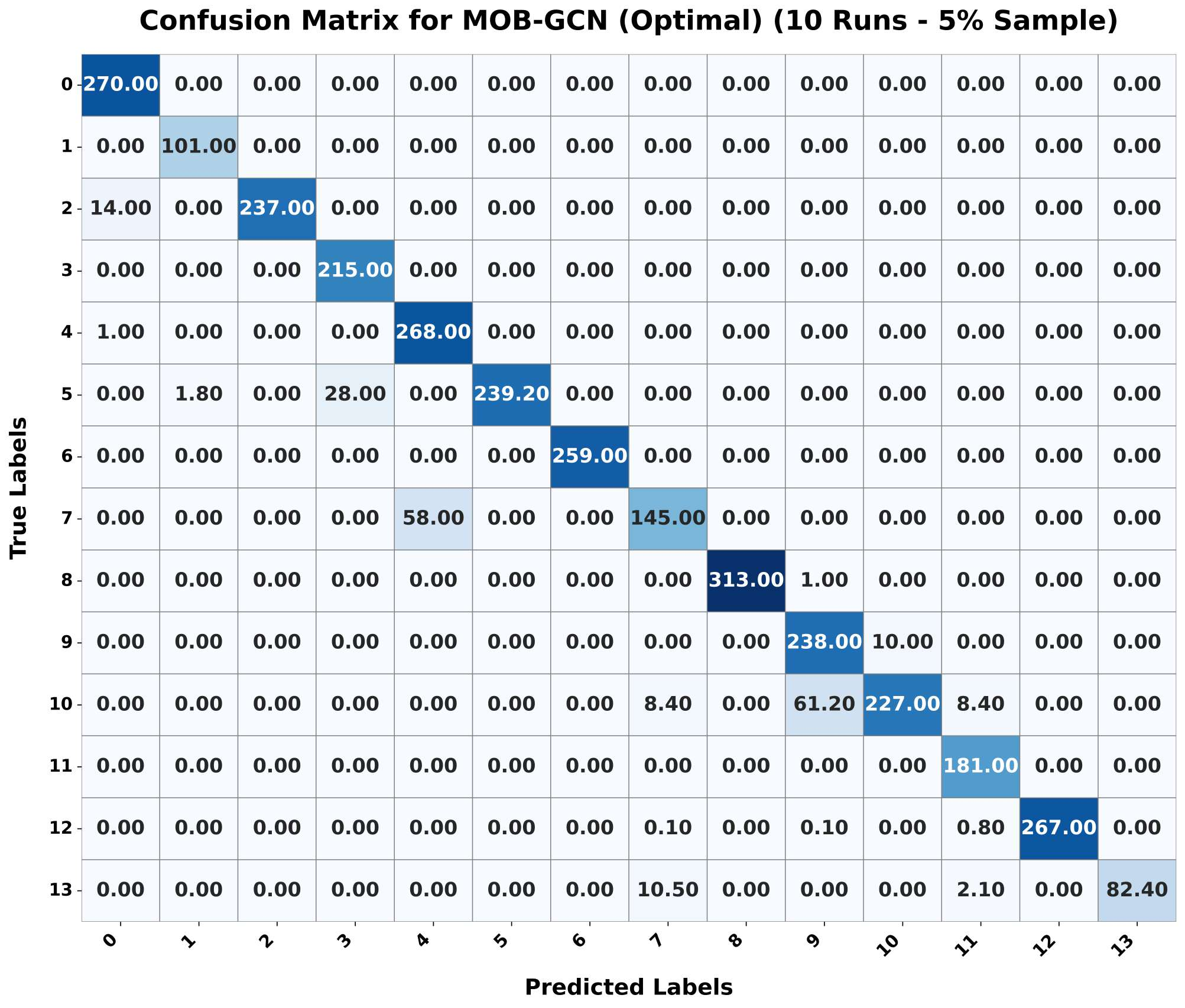} \\
\end{tabular}
\caption{BOTSWANA on 5\% Sample Data. Model comparison and Average Confusion Matrix from MOB-GCN (Optimal).}
\label{fig:botswana_metrics}
\end{figure}

\begin{figure}[h!]
\centering
\begin{tabular}{cc}
\includegraphics[width=0.5\textwidth]{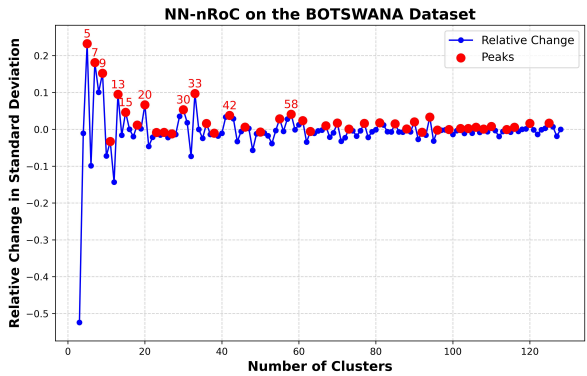} &
\includegraphics[width=0.5\textwidth]{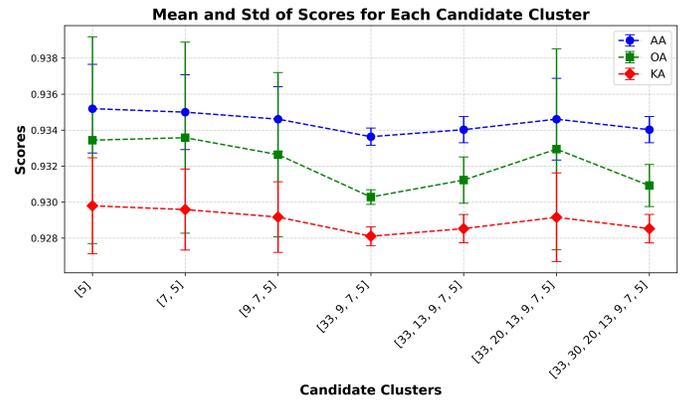} \\
\end{tabular}
\caption{BOTSWANA on 5\% Sample Data. NN-nRoC on every number of clusters on the BOTSWANA Dataset. Peaks on PAVIA’ NN-nROC are placed at 5, 7, 9, 33, 13, 20, 30, 15, 58, 42 with descending value.}
\label{fig:botswana_scales}
\end{figure}
\section{TORONTO at 5\% Sample Data} \label{sec:apx_toronto}

\begin{figure}[h!]
\centering
\begin{tabular}{cc}
\includegraphics[width=0.5\textwidth]{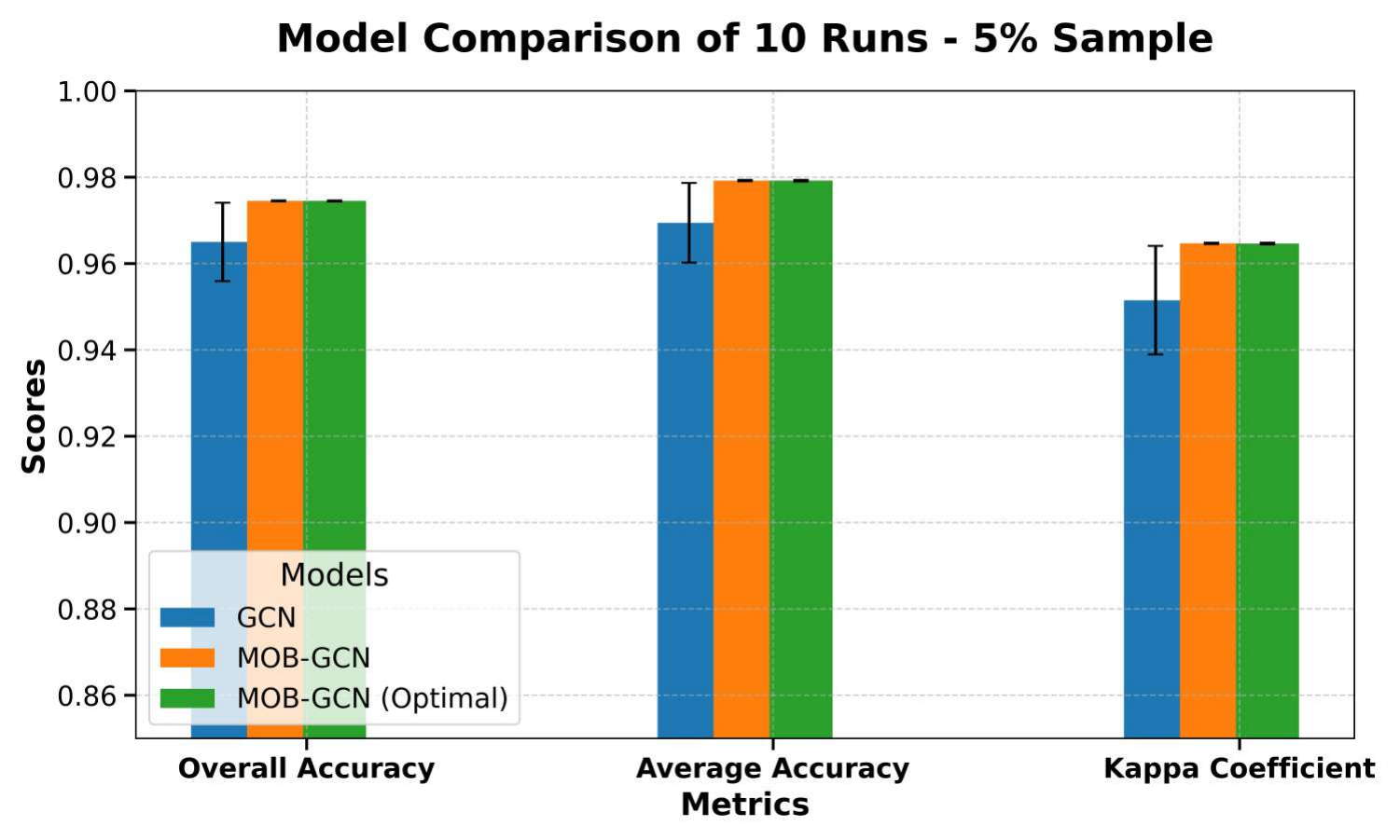} &
\includegraphics[width=0.5\textwidth]{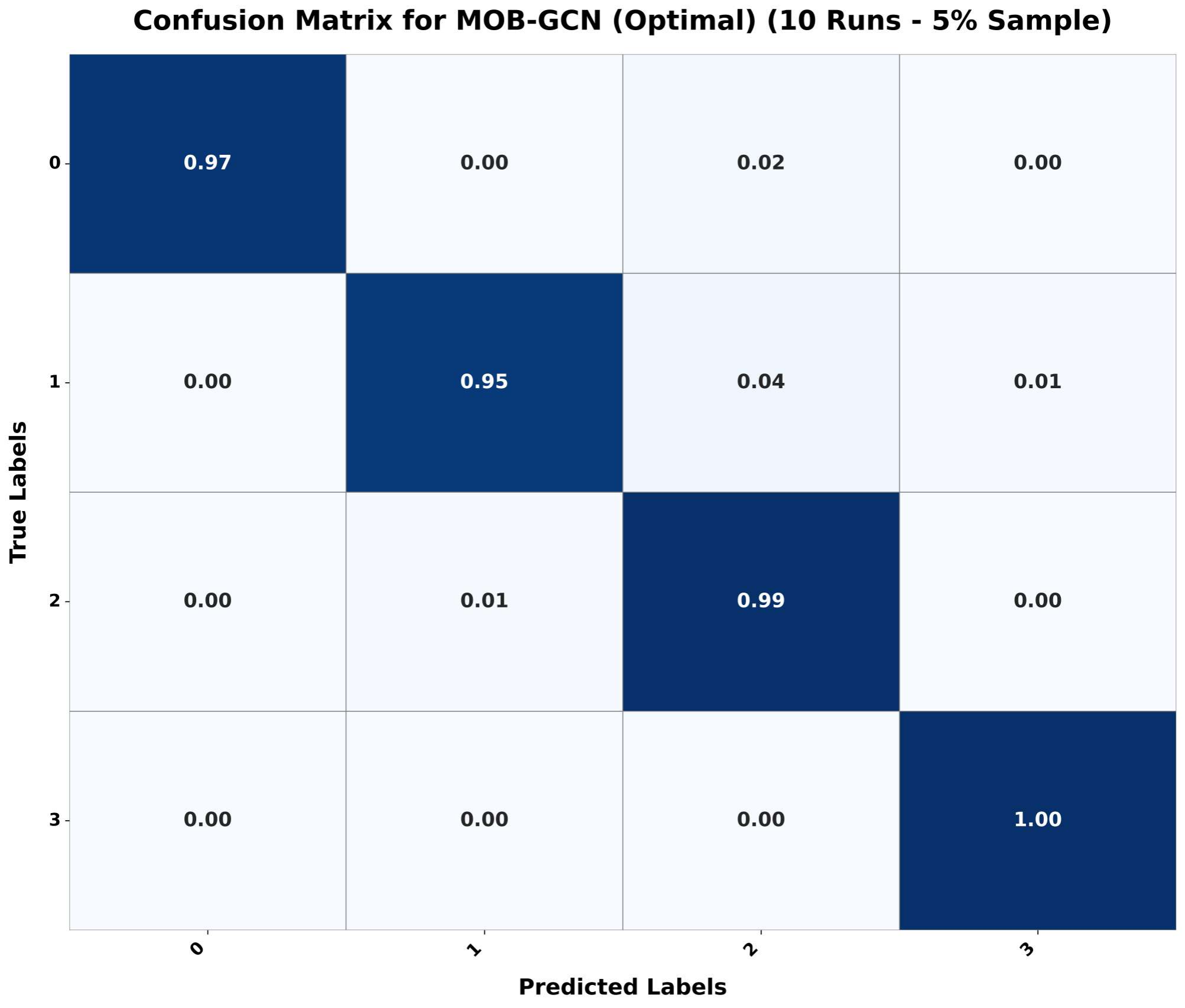} \\
\end{tabular}
\caption{TORONTO on 5\% Sample Data. Model comparison and Average Confusion Matrix from MOB-GCN (Optimal).}
\label{fig:toronto_metrics}
\end{figure}

\begin{figure}[h!]
\centering
\begin{tabular}{cc}
\includegraphics[width=0.5\textwidth]{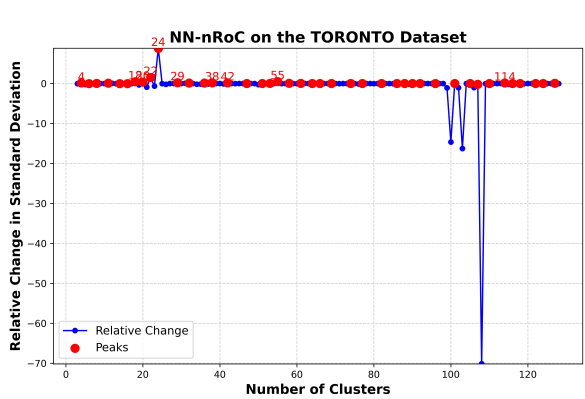} &
\includegraphics[width=0.5\textwidth]{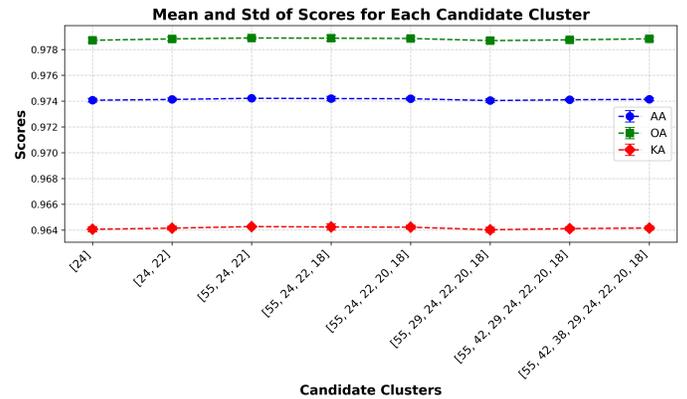} \\
\end{tabular}
\caption{TORONTO on 5\% Sample Data. NN-nRoC on every number of clusters on the TORONTO Dataset. Peaks on PAVIA’ NN-nROC are placed at 24, 22, 55, 18, 20 with descending value.}
\label{fig:toronto_scales}
\end{figure}
\section{Additional Results at 10\%, 20\% Sample Size} \label{sec:apx_results}

\begin{table}[h]
    \centering
    \caption{OA (\%) AA (\%) and Kappa (\%) of ten consecutive experiments with 10\% and 20\% sample data (* means optimal scales selected).}
    \begin{minipage}{0.49\textwidth}
\centering
\small
\begin{tabular}{|lccc|}
\hline
\multicolumn{4}{|c|}{\textbf{INDIAN}} \\
\hline
\textbf{Model} & \textbf{OA} & \textbf{AA} & \textbf{Kappa} \\
\hline
GCN & 88.70 $\pm$ 0.94 & 76.61 $\pm$ 2.95 & 87.02 $\pm$ 1.09 \\
MOB-GNN & 95.29 $\pm$ 0.43 & 94.76 $\pm$ 0.60 & 94.63 $\pm$ 0.49 \\
MOB-GNN (*) & \textbf{95.65 $\pm$ 0.18} & \textbf{95.43 $\pm$ 0.12} & \textbf{95.04 $\pm$ 0.21} \\
\hline
\multicolumn{4}{|c|}{\textbf{SALINAS}} \\
\hline
\textbf{Model} & \textbf{OA} & \textbf{AA} & \textbf{Kappa} \\
\hline
GCN & 89.19 $\pm$ 2.41 & 91.28 $\pm$ 1.77 & 87.91 $\pm$ 2.71 \\
MOB-GNN & 98.40 $\pm$ 0.23 & 98.97 $\pm$ 0.07 & 98.22 $\pm$ 0.26 \\
MOB-GNN (*) & \textbf{98.58 $\pm$ 0.21} & \textbf{99.06 $\pm$ 0.06} & \textbf{98.53 $\pm$ 0.24} \\
\hline
\multicolumn{4}{|c|}{\textbf{PAVIA}} \\
\hline
\textbf{Model} & \textbf{OA} & \textbf{AA} & \textbf{Kappa} \\
\hline
GCN & 95.70 $\pm$ 0.17 & 86.53 $\pm$ 1.22 & 93.90 $\pm$ 0.24 \\
MOB-GNN & 96.62 $\pm$ 0.23 & 91.00 $\pm$ 0.75 & 95.21 $\pm$ 0.34 \\
MOB-GNN (*) & \textbf{96.92 $\pm$ 0.24} & \textbf{92.00 $\pm$ 0.66} & \textbf{95.64 $\pm$ 0.34} \\
\hline
\multicolumn{4}{|c|}{\textbf{KENNEDY}} \\
\hline
\textbf{Model} & \textbf{OA} & \textbf{AA} & \textbf{Kappa} \\
\hline
GCN & 83.53 $\pm$ 1.48 & 71.71 $\pm$ 2.20 & 81.50 $\pm$ 1.69 \\
MOB-GNN & 92.61 $\pm$ 1.32 & 84.76 $\pm$ 1.16 & 91.85 $\pm$ 1.47 \\
MOB-GNN (*) & \textbf{93.20 $\pm$ 1.52} & \textbf{85.37 $\pm$ 1.49} & \textbf{92.41 $\pm$ 1.70} \\
\hline
\multicolumn{4}{|c|}{\textbf{BOTSWANA}} \\
\hline
\textbf{Model} & \textbf{OA} & \textbf{AA} & \textbf{Kappa} \\
\hline
GCN & 94.42 $\pm$ 2.08 & 94.46 $\pm$ 2.10 & 93.95 $\pm$ 2.25 \\
MOB-GNN & \textbf{95.64 $\pm$ 0.04} & \textbf{95.69 $\pm$ 0.04} & \textbf{95.28 $\pm$ 0.04} \\
MOB-GNN (*) & 95.62 $\pm$ 0.00 & 95.67 $\pm$ 0.00 & 95.26 $\pm$ 0.00 \\
\hline
\multicolumn{4}{|c|}{\textbf{TORONTO}} \\
\hline
\textbf{Model} & \textbf{OA} & \textbf{AA} & \textbf{Kappa} \\
\hline
GCN & 96.49 $\pm$ 0.96 & 96.83 $\pm$ 1.11 & 95.13 $\pm$ 1.33 \\
MOB-GNN & \textbf{97.61 $\pm$ 0.00} & \textbf{98.06 $\pm$ 0.00} & \textbf{96.69 $\pm$ 0.00} \\
MOB-GNN (*) & \textbf{97.61 $\pm$ 0.00} & \textbf{98.06 $\pm$ 0.00} & \textbf{96.69 $\pm$ 0.00} \\
\hline
\end{tabular}
    \end{minipage}
    \begin{minipage}{0.49\textwidth}

\centering
\small
\begin{tabular}{|lccc|}
\hline
\multicolumn{4}{|c|}{\textbf{INDIAN}} \\
\hline
\textbf{Model} & \textbf{OA} & \textbf{AA} & \textbf{Kappa} \\
\hline
GCN & 89.40 $\pm$ 1.27 & 79.85 $\pm$ 3.55 & 87.85 $\pm$ 1.45 \\
MOB-GNN & 96.60 $\pm$ 0.22 & \textbf{97.21 $\pm$ 0.14} & 96.12 $\pm$ 0.25 \\
MOB-GNN (*) & \textbf{96.74 $\pm$ 0.18} & 97.29 $\pm$ 0.11 & \textbf{96.28 $\pm$ 0.21} \\
\hline
\multicolumn{4}{|c|}{\textbf{SALINAS}} \\
\hline
\textbf{Model} & \textbf{OA} & \textbf{AA} & \textbf{Kappa} \\
\hline
GCN & 88.38 $\pm$ 2.77 & 88.78 $\pm$ 2.47 & 86.99 $\pm$ 3.05 \\
MOB-GNN & 98.44 $\pm$ 0.07 & 99.03 $\pm$ 0.11 & 98.27 $\pm$ 0.07 \\
MOB-GNN (*) & \textbf{98.91 $\pm$ 0.04} & \textbf{99.18 $\pm$ 0.02} & \textbf{98.79 $\pm$ 0.05} \\
\hline
\multicolumn{4}{|c|}{\textbf{PAVIA}} \\
\hline
\textbf{Model} & \textbf{OA} & \textbf{AA} & \textbf{Kappa} \\
\hline
GCN & 95.76 $\pm$ 0.16 & 86.18 $\pm$ 0.84 & 93.99 $\pm$ 0.22 \\
MOB-GNN & 96.55 $\pm$ 0.16 & 90.38 $\pm$ 0.43 & 95.13 $\pm$ 0.23 \\
MOB-GNN (*) & \textbf{96.81 $\pm$ 0.28} & \textbf{90.98 $\pm$ 0.68} & \textbf{95.48 $\pm$ 0.39} \\
\hline
\multicolumn{4}{|c|}{\textbf{KENNEDY}} \\
\hline
\textbf{Model} & \textbf{OA} & \textbf{AA} & \textbf{Kappa} \\
\hline
GCN & 86.02 $\pm$ 2.14 & 74.15 $\pm$ 2.78 & 84.30 $\pm$ 2.43 \\
MOB-GNN & 92.76 $\pm$ 0.23 & 83.43 $\pm$ 0.44 & 91.90 $\pm$ 0.28 \\
MOB-GNN (*) & \textbf{93.25 $\pm$ 0.70} & \textbf{84.89 $\pm$ 2.32} & \textbf{92.45 $\pm$ 0.79} \\
\hline
\multicolumn{4}{|c|}{\textbf{BOTSWANA}} \\
\hline
\textbf{Model} & \textbf{OA} & \textbf{AA} & \textbf{Kappa} \\
\hline
GCN & 93.50 $\pm$ 3.47 & 93.49 $\pm$ 3.17 & 92.96 $\pm$ 3.75 \\
MOB-GNN & \textbf{95.56 $\pm$ 0.04} & \textbf{95.54 $\pm$ 0.04} & \textbf{95.19 $\pm$ 0.05} \\
MOB-GNN (*) & 95.54 $\pm$ 0.00 & 95.52 $\pm$ 0.00 & 95.16 $\pm$ 0.00 \\
\hline
\multicolumn{4}{|c|}{\textbf{TORONTO}} \\
\hline
\textbf{Model} & \textbf{OA} & \textbf{AA} & \textbf{Kappa} \\
\hline
GCN & 95.92 $\pm$ 0.83 & 96.35 $\pm$ 0.95 & 94.35 $\pm$ 1.14 \\
MOB-GNN & \textbf{97.76 $\pm$ 0.01} & \textbf{98.25 $\pm$ 0.01} & \textbf{96.90 $\pm$ 0.01} \\
MOB-GNN (*) & 97.76 $\pm$ 0.00 & 98.24 $\pm$ 0.00 & 96.90 $\pm$ 0.00 \\
\hline
\end{tabular}
    \end{minipage}
\end{table}

\begin{table}[h]
\centering
\caption{Training and Inference Time (in seconds) Across Sample Sizes for Different Models, with 5\%, 10\% and 20\% sample data.}
\begin{tabular}{cccccccc}
\toprule
\multirow{2}{*}{Dataset} & \multirow{2}{*}{Sample Size} & \multicolumn{2}{c}{GCN} & \multicolumn{2}{c}{MGN} & \multicolumn{2}{c}{MGN (Optimal)} \\
\cmidrule(lr){3-4} \cmidrule(lr){5-6} \cmidrule(lr){7-8}
 & & Train & Infer & Train & Infer & Train & Infer \\
\midrule
\multirow{3}{*}{INDIAN} & 5\% & 1.5834 & 0.0219 & 1.8631 & 0.0187 & 3.3089 & 0.0199 \\
                        & 10\% & 1.3813 & 0.0187 & 1.8911 & 0.0203 & 3.1736 & 0.0209 \\
                        & 20\% & 1.4272 & 0.0176 & 1.9030 & 0.0183 & 3.2828 & 0.0214 \\
\midrule
\multirow{3}{*}{SALINAS} & 5\% & 1.1990 & 0.0234 & 1.6734 & 0.0242 & 3.1439 & 0.0294 \\
                         & 10\% & 1.1820 & 0.0257 & 1.7880 & 0.0281 & 3.1520 & 0.0291 \\
                         & 20\% & 1.2068 & 0.0268 & 1.7649 & 0.0269 & 3.2282 & 0.0268 \\
\midrule
\multirow{3}{*}{PAVIA} & 5\% & 1.5296 & 0.5272 & 2.0817 & 0.5667 & 3.4457 & 0.5026 \\
                       & 10\% & 1.5763 & 0.4796 & 2.0371 & 0.4778 & 3.3745 & 0.4535 \\
                       & 20\% & 1.6142 & 0.5139 & 2.1327 & 0.5044 & 3.4572 & 0.5243 \\
\bottomrule
\end{tabular}
\label{tab:training_inference_times}
\end{table}
\section{Algorithms} \label{sec:apx_algorithms}

\begin{algorithm}[ht]
\caption{Multiscale Object-based Graph Neural Network (MOB-GCN)}
\begin{algorithmic}[1]
\Require Input features $X$, adjacency matrix $A$, edge weights $W$ (optional)
\Ensure Output features

\State Initialize bottom encoder GCN
\State Initialize middle encoders and pools for each resolution level
\State Initialize final fully connected layer

\Function{Forward}{$X$, $A$, $W$}
    \State $L \gets$ BottomEncoder($X$, $A$, $W$)
    \State $L \gets \text{ReLU}(L)$
    \State allLatents $\gets [L]$
    \State $A \gets \text{ToDenseAdj}(A, W)$
    
    \State product $\gets$ None
    \For{level $= 1$ to numLevels}
        \State assign $\gets \text{GumbelSoftmax}(\text{MiddlePool}_\text{level}(L))$
        \If{level $= 1$}
            \State product $\gets$ assign
        \Else
            \State product $\gets$ product $\cdot$ assign
        \EndIf
        
        \State $X \gets \text{Normalize}(\text{assign}^T \cdot L)$
        \State $A \gets \text{Normalize}(\text{assign}^T \cdot A \cdot \text{assign})$
        
        \State $L \gets \text{ReLU}(\text{MiddleEncoder}_\text{level}(A \cdot X))$
        \State extendedLatent $\gets$ product $\cdot L$
        \State allLatents.append(extendedLatent)
    \EndFor
    
    \If{useNorm}
        \State allLatents $\gets$ [Normalize(latent) for latent in allLatents]
    \EndIf
    
    \State representation $\gets$ Concatenate(allLatents)
    \State output $\gets$ FinalFC(representation)
    
    \Return output
\EndFunction
\end{algorithmic}
\end{algorithm}

\begin{algorithm}[ht]
\caption{Training MGN with LGC Loss}
\begin{algorithmic}[1]
\Require Training dataset $\mathcal{D} = \{(X_i, A_i, W_i, y_i)\}_{i=1}^N$, number of epochs $E$, learning rate $\eta$, regularization parameter $\mu$
\Ensure Trained MGN model $\mathcal{M}$

\State Initialize MGN model $\mathcal{M}$ with parameters $\theta$
\State Initialize optimizer $\mathcal{O}$

\For{$e = 1$ to $E$}
    \For{$(X, A, W, y) \in \mathcal{D}$}
        \State $\hat{y} \gets \mathcal{M}(X, A, W; \theta)$
        
        \State // Compute supervised loss
        \If{$\mathcal{M}_{\text{train}}$ exists}
            \State $L_{\text{sup}} \gets -\sum_{i \in \mathcal{M}_{\text{train}}} y_i \log(\hat{y}_i)$
        \Else
            \State $L_{\text{sup}} \gets -\sum_{i=1}^{|y|} y_i \log(\hat{y}_i)$
        \EndIf
        
        \State // Compute smoothness regularization
        \State $d \gets \text{diag}(A\mathbf{1})$ \Comment{Node degrees}
        \State $\hat{y}_{\text{norm}} \gets D^{-1/2}\hat{y}$ \Comment{$D$ is diagonal matrix of $d$}
        \State $L_{\text{smooth}} \gets \frac{1}{|E|} \sum_{(i,j) \in E} \|\hat{y}_{\text{norm},i} - \hat{y}_{\text{norm},j}\|_2^2$
        
        \State // Compute total loss
        \State $L_{\text{total}} \gets L_{\text{sup}} + \mu L_{\text{smooth}}$
        
        \State // Update model parameters
        \State $\theta \gets \theta - \eta \nabla_\theta L_{\text{total}}$
    \EndFor
\EndFor

\Return Trained MGN model $\mathcal{M}$
\end{algorithmic}
\end{algorithm}

\end{document}